\pdfoutput=1

\documentclass[11pt, table]{article}

\usepackage[]{acl}

\usepackage{times}
\usepackage{latexsym}
\usepackage{xcolor}
\usepackage[T1]{fontenc}

\usepackage[utf8]{inputenc}

\usepackage{microtype}

\usepackage{inconsolata}

\usepackage{multirow}
\usepackage{booktabs}
\usepackage{rotating}
\usepackage{footnote} 

\title{MALAMUTE: A Multilingual, Highly-granular, Template-free, Education-based Probing Dataset}

\author{
\textnormal{$^\nabla$Sagi Shaier, $^\nabla$George Arthur Baker$^*$, $^\nabla$Chiranthan Sridhar\thanks{These authors contributed equally to this work.}} \\
\textnormal{$^\dag$Lawrence E Hunter, Katharina von der Wense$^{\nabla\diamondsuit}$} \\
$^\nabla$University of Colorado Boulder \\
$^\dag$University of Chicago, Department of Pediatrics \\
$^\diamondsuit$Johannes Gutenberg University Mainz \\
E-mail: \{sagi.shaier, george.baker, chiranthan.sridhar, katharina.kann\}@colorado.edu
}

\begin{document}
\maketitle
\begin{abstract}
Language models (LMs) have excelled in various broad domains. However, to ensure their safe and effective integration into real-world educational settings, they must demonstrate proficiency in specific, granular areas of knowledge. Existing cloze-style benchmarks, commonly used to evaluate LMs' knowledge, have three major limitations. They: 1) do not cover the educational domain; 2) typically focus on low-complexity, generic knowledge or broad domains, which do not adequately assess the models' knowledge in specific subjects; and 3) often rely on templates that can bias model predictions. Here, we introduce MALAMUTE, a multilingual, template-free, and highly granular probing dataset comprising expert-written, peer-reviewed probes from 71 university-level textbooks across three languages (English, Spanish, and Polish). MALAMUTE is the first education-based cloze-style dataset. It covers eight domains, each with up to 14 subdomains, further broken down into concepts and concept-based prompts, totaling 33,361 university curriculum concepts and 116,887 prompts. MALAMUTE's fine granularity, educational focus, and inclusion of both sentence-level and paragraph-level prompts make it an ideal tool for evaluating LMs' course-related knowledge. Our evaluation of masked and causal LMs on MALAMUTE shows that despite overall proficiency, they have significant gaps in knowledge when examined closely on specific subjects, hindering their safe use in classrooms and underscoring the need for further development. Code and data can be found here \href{https://github.com/Shaier/MALAMUTE}{https://github.com/Shaier/MALAMUTE}
\end{abstract}

\begin{figure}[ht]
  \centering
  \includegraphics[width=.45\textwidth]{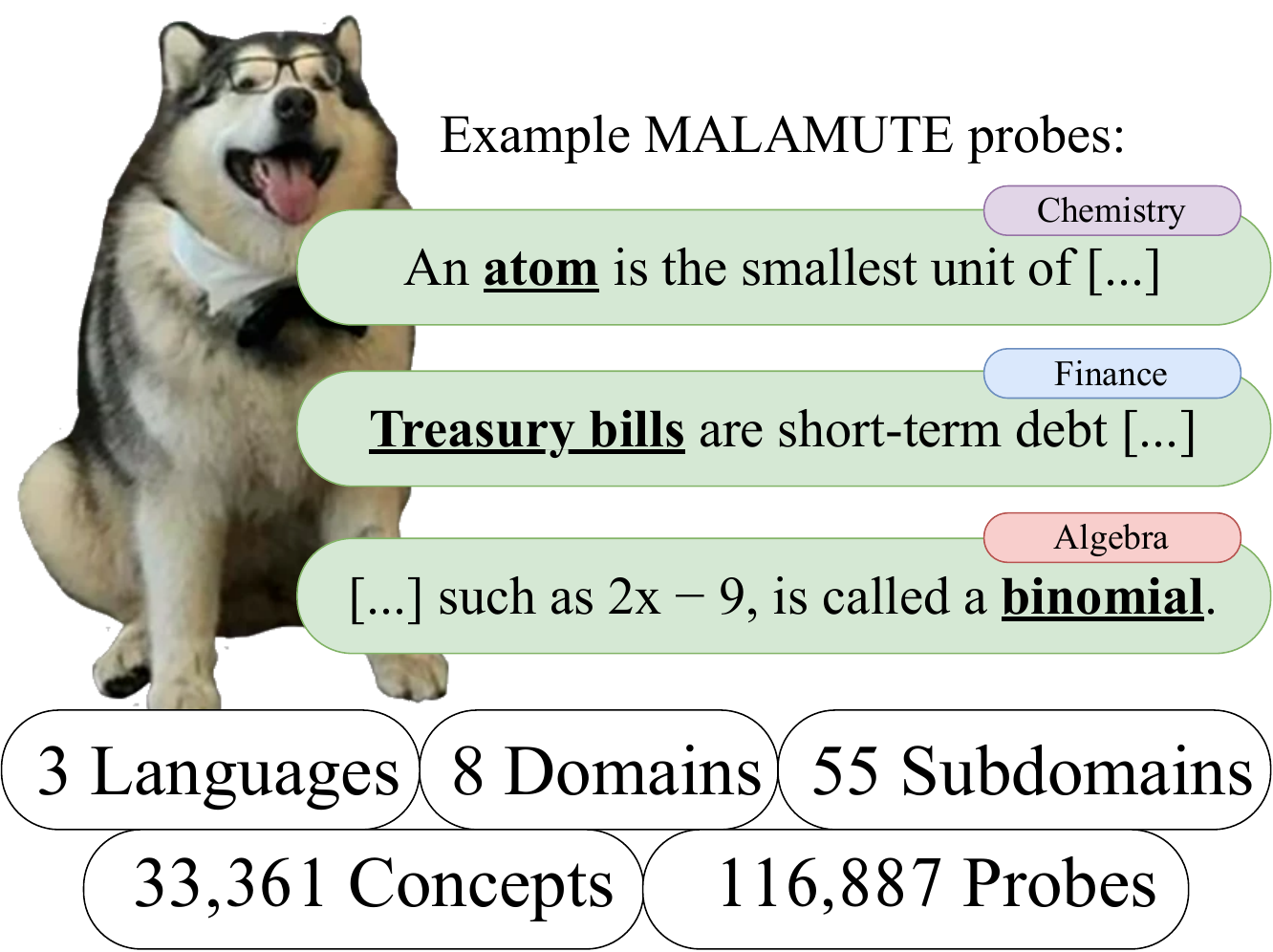}
  \caption{
  MALAMUTE: a highly granular cloze-style dataset for educational knowledge probing. This dataset assesses language models' knowledge across three languages with fine-grained detail. \textbf{Bold strings} indicate predictable text. Example prompts shown for 3 out of the 55 subdomains.
  }
  \label{malamute_fp}
\end{figure}

\section{Introduction}
Language models (LMs) have shown remarkable abilities in various domains \cite{openai_2023, openai2023gpt4, touvron2023llama}. However, to ensure their safe and effective integration into real-world educational settings, they must demonstrate proficiency in specific, granular areas of knowledge \cite{xu-etal-2020-multi}. For instance, a LM may excel in general knowledge of mathematics, but struggle with the nuances of calculus or algebra. Hence, we need precise and targeted evaluation methods for LMs' knowledge in these critical areas.

The need for such evaluations is pressing, as LMs are increasingly being considered for use in classrooms, where their performance can have a direct and profound impact on student learning outcomes \cite{jovst2024impact, kumar2023impact}. Moreover, LMs are being explored for a range of educational applications, including adaptive learning \cite{weijers-etal-2024-quantifying}, intelligent tutoring systems \cite{nye2023generative}, and automated grading \cite{xie2024grade}. In order to ensure that these systems are both effective and safe, we must develop evaluation methods that can accurately assess their knowledge in specific, curriculum-aligned areas. 

Currently, evaluating LMs' knowledge is a significant research focus.
Two prominent evaluation methods are question answering (QA) \cite{shaier-etal-2024-desiderata, li-etal-2023-large, xie2024adaptive} and cloze-style (CS) probing \cite{petroni-etal-2019-language, sung-etal-2021-language}. CS probing, in particular, has gained popularity, where models receive a prompt with a masked portion, such as “Dante was born in [MASK],” and are tasked with predicting the masked string. This approach measures models' lower bound of knowledge \cite{petroni-etal-2019-language}. 

However, existing CS benchmarks have limitations. Firstly, there is a notable lack of CS datasets that are specifically designed for educational settings. Secondly, the prompts in existing datasets typically focus on low-complexity, generic knowledge or broad domains, which do not adequately capture models' knowledge in specific subjects \cite{petroni-etal-2019-language, keleg-magdy-2023-dlama, kassner2020negated, kassner2020pretrained, dhingra-etal-2022-time}. Furthermore, most existing cloze-style probing datasets rely on expert-written templates, which have been shown to bias model predictions \cite{cao-etal-2021-knowledgeable, petroni-etal-2019-language, sung-etal-2021-language, shaier-etal-2024-comparing}.

To address these limitations, we introduce MALAMUTE, a MultilinguAl, highLy grAnular, teMplate-free, edUcation-based probing daTasEt. Derived from 71 university-level textbooks across three languages (English, Spanish, and Polish), MALAMUTE covers eight domains, each with up to fourteen subdomains, broken down into individual curriculum concepts and concept-centric prompts. With 33,361 curriculum concepts and 116,887 expert-written, peer-reviewed CS prompts, MALAMUTE offers an unprecedented level of granularity and alignment with educational curricula, making it an ideal tool for evaluating LMs' knowledge in educational contexts. See Figure \ref{malamute_fp} for examples. MALAMUTE is also the first probing dataset to include both sentence-level and paragraph-level prompts, which provides a more comprehensive evaluation of LMs' knowledge, establishing a new higher lower-bound of knowledge for masked LMs (MLMs). Lastly, MALAMUTE also comes with a number of parallel domains, subdomains, and concepts across multiple languages.

We evaluate MALAMUTE on 8 causal LMs (CLMs) and 5 MLMs and highlight the need for granular datasets: despite many models' overall proficiency on the dataset, they have significant gaps in knowledge when examined closely on specific subjects, hindering their safe use in classrooms and underscoring the need for further development.

\section{Related Work}
\subsection{Model's Parametric Knowledge}
LMs have been shown to contain a substantial amount of factual knowledge within their parameters, as demonstrated by \citet{petroni-etal-2019-language, sung-etal-2021-language, roberts-etal-2020-much}. This embedded knowledge enables LMs to perform tasks requiring extensive information, like open-domain question answering, without needing external databases or knowledge sources \cite{roberts-etal-2020-much}. As models' size increase, their capacity to store and retrieve vast amounts of information increases significantly \cite{openai_2023, openai2023gpt4}, which also allow them to encompass a wide range of knowledge types, including factual data \cite{petroni-etal-2019-language, shaier-etal-2023-stochastic}, linguistic nuances \cite{zhang-choi-2021-situatedqa}, cultural references \cite{zhou2024does, keleg-magdy-2023-dlama}, and domain-specific information \cite{sung-etal-2021-language}. 

Evaluating the amount and types of knowledge within these models is crucial, as it ensures that they can be effectively and reliably applied to various tasks. This assessment helps identify gaps and biases, informing improvements in training and deployment \cite{shaier-etal-2023-emerging}. Moreover, understanding models' knowledge scope is crucial for ethical considerations, such as preventing misinformation and ensuring factual language generation \cite{wang2023surveyfactualitylargelanguage, shaier-etal-2024-adaptive}.

\subsection{Knowledge Probing}
Knowledge probing is an approach to assess knowledge implicitly embedded in LMs' weights. Various methods are utilized to evaluate this parametric knowledge, including closed-book QA \cite{shaier-etal-2024-desiderata, li-etal-2023-large, xie2024adaptive}  and CS prompts \cite{petroni-etal-2019-language, sung-etal-2021-language}. Closed-book QA tests models without access to external resources, which can further be separated into those that use multiple-choice QA, or free-form QA, while CS prompts require completion of missing segments within a given context. Evaluators may further employ template-based or template-free probes to gauge comprehension, where the latter provide a more accurate assessment of LMs knowledge \cite{shaier-etal-2024-comparing}.

\subsection{Cloze-style Knowledge Probing}
CS knowledge probing has experienced a surge in popularity in recent years, resulting in the development of over 80 distinct datasets to date, spanning multiple languages and domains. CS probing involves presenting models with prompts that have masked portions, such as “Dante was born in [MASK]”, and tasking them with predicting the correct completion \cite{petroni-etal-2019-language, petroni2020context, zhou2024does, talmor2020olmpics,sung-etal-2021-language, kassner2020negated, kassner2020pretrained, dhingra-etal-2022-time}. This method is used to measure models' lower bound of knowledge.  \cite{petroni-etal-2019-language, jiang2020know}. 

While existing datasets span diverse domains, such as law \cite{chalkidis-etal-2023-lexfiles}, geography \cite{10.1145/3589132.3625625}, biomedicine \cite{sung-etal-2021-language, shaier-etal-2024-comparing, meng-etal-2022-rewire}, and general knowledge \cite{petroni-etal-2019-language, cao2021knowledgeable, he2024language}, they have several limitations. Firstly, the prompts in existing datasets typically focus on low-complexity, generic knowledge or broad domains, which do not adequately capture models' knowledge in specific subjects. Secondly, most existing cloze-style probing datasets rely on expert-written templates, which have been shown to bias model predictions. Furthermore, there is a notable lack of CS datasets that are specifically designed for educational settings. 

\subsection{Educational Knowledge Probing}
As LMs are increasingly being integrated into classrooms \cite{kazemitabaar2024codeaid, cao2023designing}, it is essential to have a reliable educational knowledge-probing dataset to assess their performance, which can significantly impact student learning outcomes. However, current educational probing datasets have several limitations.

One major concern is that some datasets rely on expert-written templates \cite{koto2024arabicmmlu}, which can introduce bias into model predictions. Furthermore, these datasets usually have a limited scope, focusing on high-level subjects (such as American Government and U.S. history \cite{ciosici-etal-2021-perhaps}) rather than providing the fine-grained, concept-level details that are necessary for in-depth knowledge. This is particularly problematic in education, where many concepts comprise various subdomains that require comprehensive assessment.

Another limitation is that these datasets have a limited number of prompts, restricting their ability to comprehensively evaluate LMs. Additionally, they are not based on CS prompts, instead relying on either multiple-choice QA or closed-book settings. However, multiple-choice QA is a significantly less challenging task due to the limited number of options that can skew knowledge assessment results and can be subject to models' biases towards lettered answer options in causal setups \cite{zheng2023large}. Closed-book settings, on the other hand, are not suitable for MLMs without additional fine-tuning or mask tokens, which can lead to grammatical issues. Moreover, evaluating free-form text generation often relies on human assessment to fully capture its correctness, as automatic metrics may not fully capture the nuances of the generated output, which can be time-consuming and subjective. Hence, there is a pressing need for a more comprehensive and nuanced CS educational knowledge-probing dataset. A summary of such probing datasets can be found in Table \ref{tab:datasets}.

\begin{table*}[t]
\centering
\setlength{\tabcolsep}{1.5pt}
\resizebox{\textwidth}{!}{%
\begin{tabular}{lccccccc}
\toprule
\textbf{Dataset} & \textbf{Template-free} & \textbf{Educational Focus} & \textbf{Cloze-style} & \textbf{Domains} & \textbf{Prompts} & \textbf{Languages} & \textbf{Domain Granularity} \\
\midrule
LAMA \cite{petroni-etal-2019-language} & \cellcolor{green!25}Both & \cellcolor{red!25}No & \cellcolor{green!25}Yes & 1* & 51k & 1 & Low \\
IndicGLUE$^\dagger$ \cite{kakwani-etal-2020-indicnlpsuite} & \cellcolor{green!25}Yes & \cellcolor{red!25}No & \cellcolor{green!25}Yes & 1* & 239k & 11 & Low \\
mLAMA \cite{kassner-etal-2021-multilingual} & \cellcolor{red!25}No & \cellcolor{red!25}No & \cellcolor{green!25}Yes & 1* & 855k & 53 & Low \\
MedLAMA \cite{meng-etal-2022-rewire} & \cellcolor{red!25}No & \cellcolor{red!25}No & \cellcolor{green!25}Yes & 1 & 19k & 1 & Low \\
BioLAMA \cite{sung-etal-2021-language} & \cellcolor{red!25}No & \cellcolor{red!25}No & \cellcolor{green!25}Yes & 1 & 49k & 1 & Low \\
LegalLAMA \cite{chalkidis-etal-2023-lexfiles} & \cellcolor{green!25}Yes & \cellcolor{red!25}No & \cellcolor{green!25}Yes & 1 & 27k & 1 & Low \\
NumerSense \cite{lin-etal-2020-birds} & \cellcolor{red!25}No & \cellcolor{red!25}No & \cellcolor{green!25}Yes & 8 & 3k & 1 & Medium \\
LEFT \cite{ciosici-etal-2021-perhaps} & \cellcolor{green!25}Yes & \cellcolor{green!25}Yes & \cellcolor{red!25}No & 2 & 1k & 1 & Medium \\
EXAMS$^\dagger$ \cite{hardalov-etal-2020-exams} & \cellcolor{green!25}Yes & \cellcolor{green!25}Yes & \cellcolor{red!25}No & 3 & 14k & 16 & High \\
ArabicMMLU$^\dagger$ \cite{koto2024arabicmmlu} & \cellcolor{red!25}No & \cellcolor{green!25}Yes & \cellcolor{red!25}No & 5 & 15k & 2 & High \\
\hline
MALAMUTE & \cellcolor{green!25}Yes & \cellcolor{green!25}Yes & \cellcolor{green!25}Yes & \textbf{8} & 116k & 3 & Very High \\
\bottomrule
\end{tabular}%
}
\caption{Comparison of recent knowledge probing datasets. 
Low domain granularity means a dataset is not divided into domains, though it may have multiple tasks or sources. Medium granularity involves several domains without subdomains. High granularity includes domains and subdomains, while very high granularity adds a further division into concepts. * These datasets cover broad general knowledge and do not correspond to a specific domain. $^\dagger$ multiple-choice QA.}
\label{tab:datasets}
\end{table*}

\section{MALAMUTE}
\begin{table*}[h]
\centering
\tiny
\begin{tabular}{llll}
\hline
\textbf{Domain} & \textbf{Subdomain} & \textbf{Prompt} & \textbf{Concept}\\
\hline
\multirow{5}{*} 
        & Prealgebra & We call -a the \textbf{[MASK]} of a. & additive inverse \\
        
        & Elementary Algebra & In \textbf{[MASK]}, we express x>3 as (3,$\infty$). & interval notation \\
        
        & Intermediate Algebra & If we reverse the x and y in the function and then solve for y, we get our \textbf{[MASK]}. & inverse function \\
        
        & College Algebra & A polynomial containing two terms, such as $2x-9$, is called a \textbf{[MASK]}. & binomial \\
        
        & Trigonometry & An \textbf{[MASK]} is an even root of a negative number. & imaginary number \\
        
        & Precalculus $\textcolor{purple}\star$ & The \textbf{[MASK]} can be used to determine whether a graph represents a function. & vertical line test \\
        
        Math& Calculus I $\textcolor{purple}\star$ & If f' is decreasing over I, we say f is \textbf{[MASK]} over I. & concave down \\
        
        & Calculus II $\textcolor{purple}\star$ & The net change theorem considers the integral of a \textbf{[MASK]}. & rate of change \\
    
        & Calculus III $\textcolor{purple}\star$ & The nonzero vectors u and v are \textbf{[MASK]} if and only if u·v=0. & orthogonal vectors \\

        & Introductory Statistics & A \textbf{[MASK]} is the number of times a value of the data occurs. & frequency \\

        & Business Statistics $\textcolor{purple}\star$ & The \textbf{[MASK]} is the most frequent value. & mode \\
        
        & Statistics $\textcolor{purple}\star$ & A \textbf{[MASK]} provides a way of portraying data that can facilitate calculating probabilities. & two-way table \\
        
        & Contemporary Math & A \textbf{[MASK]} is any voting method that satisfies the Condorcet criterion. & Condorcet method \\
        \hline

\multirow{5}{*} 
        & Concepts of Biology & All the individuals of a species living within a specific area are collectively called a \textbf{[MASK]}. & population \\
        
        & Biology & There are very few living species of \textbf{[MASK]}: the platypus and four species of echidnas [...] & monotremes \\

        & Microbiology &The reactions of nitrogen fixation occur in specialized cells called \textbf{[MASK]}. & heterocysts \\
        
        & Chemistry Atoms First & The number of protons in the nucleus of an atom is its \textbf{[MASK]}. & atomic number \\
        
        & Chemistry $\textcolor{purple}\star$ & An \textbf{[MASK]} is the smallest unit of an element that can participate in a chemical change. & atom \\
        
        & Organic Chemistry & \textbf{[MASK]} complexed with protein provides the physical makeup of the ribosomes. & Ribosomal RNA \\
    
        Science& Anatomy & The psoas major and iliacus make up the \textbf{[MASK]}. & iliopsoas group \\
        
        & Astronomy & \textbf{[MASK]}, the closest planet to the Sun [...] & Mercury \\

        & College Physics & Physics as it developed from the Renaissance to the end of the 19th century is called \textbf{[MASK]}. & classical physics \\

        & Physics & The boiling point of water is 100 °C for the Celsius scale, and its unit is the \textbf{[MASK]}. & degree Celsius \\
        
        & Physics I $\textcolor{red}\diamond$ $\textcolor{purple}\star$ & For elliptical orbits, the point of closest approach of a planet to the Sun is called the \textbf{[MASK]}. & perihelion \\

        & Physics II $\textcolor{red}\diamond$ $\textcolor{purple}\star$ & A special type of potential difference is known as [MASK]. & electromotive force \\

        & Physics III $\textcolor{red}\diamond$ $\textcolor{purple}\star$ & Electroweak theory unifies the theory of \textbf{[MASK]}[...] & quantum electrodynamics \\
        \hline
        
\multirow{5}{*} & U.S. History & The end of the Civil War saw the beginning of the \textbf{[MASK]} era [...] & Reconstruction \\
        & Writing & \textbf{[MASK]} : reference to the source of information used in a writer’s research. & Citation \\
    Humanities& Philosophy & German philosopher Friedrich \textbf{[MASK]} famously declared that “God is dead” & Nietzsche \\
        & World History I & Conflict between \textbf{[MASK]} and the Sui began in the 590s and lasted for decades. & Goguryeo \\
        & World History II & In 1864, socialists founded the International Workingmen’s Association (IWA) in \textbf{[MASK]}. & London \\
        \hline

\multirow{5}{*} 
        & Introduction to Business & \textbf{[MASK]} are the individuals or groups to whom a business has a responsibility. & Stakeholders \\

        & Financial Accounting & The \textbf{[MASK]} of cost allocation assumes that the last units purchased are the first units sold. & last-in, first-out method \\

        & Managerial Accounting & The person overseeing all of the accounting and finance concerns is the \textbf{[MASK]}. & Chief Financial Officer \\
        
        & Business Ethics & Acting with \textbf{[MASK]} means we adhere strongly to a code of ethics[...] & integrity \\
        
        & Organizational Behavior & Once acceptable behavioral criteria have been specified, a \textbf{[MASK]} can be done. & performance audit \\
        
        & Finance & \textbf{[MASK]}, are short-term debt instruments issued by the federal government. & Treasury bills \\
        
        & Business Law & Another remedy is \textbf{[MASK]}, which would terminate the right of a partnership to exist. & dissolution \\
        
        Business& Intellectual Property & The fundamental and overriding requirement for a trademark is \textbf{[MASK]}. & distinctiveness \\
    
        & Marketing & \textbf{[MASK]} happens when those being observed aren’t aware that they are being watched. & Unobtrusive observation \\
        
        & Management & \textbf{[MASK]} are desired goals, objectives, or end states that individuals wish to pursue. & Terminal values \\
        
        & Economics & We call a firm’s first stock sale to the public an \textbf{[MASK]}. & initial public offering \\
        
        & Macroeconomics $\textcolor{red}\diamond$ & The simplest example of a rate of return is the \textbf{[MASK]}. & interest rate \\
        
        & Microeconomics $\textcolor{red}\diamond$ & In the case of \textbf{[MASK]}, one firm produces all of the output in a market. & monopoly \\
        
        & Entrepeneurship & A \textbf{[MASK]} is a company that does not allow members of the investing public to own stock. & privately held corporation \\
        \hline

\multirow{1}{*} 
    Nursing    & Nutrition & Any change in sodium and \textbf{[MASK]} can dramatically affect the cells of the brain. & water balance \\
        \hline

\multirow{1}{*}
     & Python Programming & \textbf{[MASK]} is the task of arranging elements in a sequence in ascending or descending order. & Sorting \\

    Computer Science& Workplace Software & The second most popular operating system to emerge during this time was the \textbf{[MASK]}, & Android operating system \\

    &  & first developed in 2005 and later acquired by Google. & \\

        \hline

\multirow{5}{*} & American Government & The \textbf{[MASK]} limits the ability of the government to control or restrict religious practices. & free exercise clause \\
        & Anthropology & Another group, called “untouchables” or \textbf{[MASK]}, are outside the scheme of varnas. & dalits \\
    Social Sciences    & Political Science & The Greek philosopher \textbf{[MASK]} argued that humans were “political animals” [...] & Aristotle \\
        & Psychology $\textcolor{red}\diamond$ & The behavior caused by the conditioned stimulus is called the \textbf{[MASK]}. & conditioned response \\
        & Sociology & The power in an \textbf{[MASK]} is held by a small, elite group. & oligarchy \\
        \hline
\multirow{1}{*}
    & College Success & If we rely too heavily on assumptions, we may be buying into \textbf{[MASK]}, or generalizations & Stereotypes \\

    College Success& College Success Concise & One particular studying technique is called \textbf{[MASK]}, which calls for students to mix [...] & interleaving
    \\
        \hline
\end{tabular}
\caption{\label{citation-guide}
Examples of probes from each of the English textbooks.
Books that are also in Polish are marked with $\textcolor{red}\diamond$, where books that are also in Spanish are marked with $\textcolor{purple}\star$. }
\end{table*}
In this section, we introduce MALAMUTE, the first CS educational knowledge probing dataset. We also describe our systematic and scalable approach to generating high-quality probing prompts directly from reputable academic sources. See Table \ref{citation-guide} for prompt examples from each of the 8 domains and 55 subdomains in MALAMUTE.

\subsection{Data-source: OpenStax}
OpenStax \cite{openstax}, a project led by Rice University, is dedicated to creating and distributing high-quality, peer-reviewed, and open-source textbooks for popular university-level general education courses. These textbooks are made available under a Creative Commons license, allowing for free use and adaptation.

The OpenStax library currently contains an impressive collection of 70 English textbooks, 11 Spanish textbooks, and 6 Polish textbooks. All textbooks can be accessed in multiple formats: print, PDF, and interactive web pages. Moreover, the library is continuously expanding, with new subjects, textbook improvements, and language translations being added regularly. \textbf{This ongoing development ensures that our dataset will remain extensible and adaptable in the future.}

The textbooks are structured in a hierarchical system, with eight main categories (or “\textit{domains}”): Business, College Success, Computer Science, Humanities, Math, Nursing, Science, and Social Sciences. Each domain is further divided into subjects, which we refer to as “\textit{subdomains}”. These subdomains roughly correspond to the content of a single general education college course. Finally, each subdomain contains numerous \textit{concepts}, which are linked to specific sections within the relevant textbooks where those concepts are discussed.

\subsection{Dataset Creation}
\subsubsection{Data Extraction}
We collect MALAMUTE by first retrieving a list of textbook URLs from the OpenStax sitemap. We then drop remedial instances such as earlier editions and high-school level textbooks, as their contents are largely redundant, and compile a list of index pages. Next, we iterate over the vocabulary terms in the index pages and scrape the contents of the pages associated with the hypertext reference attributes. For each page, we save any parent object of a span with the attribute “data-type” equal to “term”, which generally corresponds to the surrounding paragraph. We exclude rare cases that result in prompts consisting of only the concept itself, primarily occurring in tables, which would be inappropriate for cloze-style prompts. \textbf{The term spans we select are chosen by domain-expert authors of the textbooks and undergo multiple rounds of peer review, with educational value and informativeness as the primary criteria}. This makes them ideal candidates for knowledge probing prompts.

\subsubsection{Cloze-style Prompts Creation}
While existing benchmark largely focus on sentence-level prompts, MALAMUTE is the first CS dataset that is composed of both \textit{sentence-level} and \textit{paragraph-level} prompts.

\paragraph{Paragraph-level Prompts} We take each unique term in each paragraph and replace every occurrence of that term with the “[MASK]” string. We store the original term as a label and hide all other occurrences of the term in the same paragraph by replacing them with the “[HIDDEN]” string. This results in prompts that provide more contextual information about the term.

\paragraph{Sentence-level Prompts} We split each paragraph into individual sentences using the NLTK library. Then, we select only the sentences that contain the term we are interested in. This results in a more challenging prompt, as it focuses on a single sentence and provides less contextual information.

\subsubsection{Filtration}
To ensure the quality of our prompts, we developed a comprehensive filtration methodology, which is described in detail in Appendix \ref{filtration}. This methodology involves a combination of regular expressions and POS tagging to remove prompts that do not meet our quality standards, as determined through a thorough manual review of the prompt set.

\subsection{Quality Control}
To ensure the quality of our generated prompts, we conducted a thorough quality control process, which is described in detail in Appendix \ref{quality_control}. In brief, 
Graduate student annotators with academic backgrounds spanning MALAMUTE's diverse subject areas,
as evident from their academic transcripts, manually reviewed a representative sample of 495 English prompts. The results showed that the prompts demonstrated a high level of quality, with $97.6\%$ of paragraph-level prompts and $99.1\%$ of sentence-level prompts meeting grammatical standards, and $92.9\%$ and $93.9\%$, respectively, effectively conveying the intended concept. This rigorous quality control process verified that the generated prompts meet the desired standards, providing a robust foundation for evaluating the performance of LLMs.

\subsection{MALAMUTE Statistics}
Our methodology yields a substantial dataset of 116,887 unique prompts, covering 33,361 distinct concepts across 71 textbooks and three languages. The language distribution of the prompts is: 100,258 prompts are in English, 8,147 in Polish, and 8,482 in Spanish.
Domain-level statistics are provided in Table \ref{tab:domain_stats}.

\begin{table*}[t]
\centering
\small
\setlength{\tabcolsep}{1.5pt}
\begin{tabular}{c lcccccc}
\toprule
 & \textbf{Domain} & \textbf{\#Books} & \textbf{\#Prompts} & \textbf{\#Concepts} & \textbf{Words (Pg.)} & \textbf{Words (Sent.)}  \\
\cmidrule{1-7}

\multirow{8}{*}{\begin{sideways} EN\end{sideways}}

& Business & 14 & 18,373 & 7,923 & 105.27 & 22.84 \\
& College Success & 2 & 206 & 115 & 104.09 & 18.60 \\
& Computer Science & 2 & 3,682 & 1,447 & 93.75 & 18.44 \\
& Humanities & 5 & 18,622 & 6,342 & 114.47 & 24.78 \\
& Math & 13 & 7,103 & 3,384 & 55.17 & 18.44 \\
& Nursing & 1 & 3,934 & 1,567 & 96.85 & 23.32 \\
& Science & 13 & 35,095 & 15,602 & 116.55 & 21.87 \\
& Social Sciences & 5 & 13,696 & 5,792 & 139.82 & 26.31 \\

\cmidrule{1-7}

\multirow{3}{*}{\begin{sideways} PL \end{sideways}}
& Science & 3 & 3,428 & 1,684 & 101.33 & 16.60 \\
& Business & 2 & 1,660 & 771 & 115.07 & 20.25 \\
& Social Sciences & 1 & 3,059 & 1,481 & 93.74 & 17.19 \\
\cmidrule{1-7}

\multirow{2}{*}{\begin{sideways} ES \end{sideways}}
& Math & 6 & 3,481 & 1,512 & 61.14 & 19.84 \\
& Science & 4 & 5,001 & 2,388 & 128.64 & 23.45 \\

\bottomrule

\end{tabular}
\caption{Statistics of MALAMUTE. Pg and Sent refer to the average \#words in paragraph and sentence level.}
\label{tab:domain_stats}
\end{table*}





\section{Experiments}
\subsection{Models}

We evaluate a set of five MLMs: BERT-base \cite{devlin-etal-2019-bert}, DistilBERT-base \cite{sanh2020distilbert}, and SciBERT, a variant pretrained on scientific text \cite{beltagy2019scibert}. Additionally, we include two multilingual models, XLM-RoBERTa-large \cite{conneau-etal-2020-unsupervised} and mBERT-base \cite{devlin-etal-2019-bert}, which are specifically designed to handle a wide range of languages.

We also assess eight CLMs: GPT-4o, GPT-4o-mini, and GPT-4-turbo \cite{achiam2023gpt}, which are state-of-the-art generative models; Llama-3.1-405B-Instruct \cite{dubey2024llama}, a powerful model designed for instruction-based tasks; Bloom-7B \cite{muennighoff2023crosslingual} and Mistral-7B-Instruct \cite{jiang2023mistral}, which are trained for cross-lingual tasks; and Llama-2-7B-Chat \cite{touvron2023llama} and Llama-3-8B-Instruct \cite{Llama3}, both pretrained on multilingual data, with a focus on both general and scientific domains.

This comprehensive evaluation includes a range of models from both the MLM and CLM categories, pretrained on general and scientific data to ensure broad coverage of linguistic and domain-specific tasks.

\subsection{Evaluation Metric}
\label{eval}
\textbf{MLM Evaluation:} Unlike previous work, which mostly focuses on single-token mask prediction \cite{zhong-etal-2021-factual, petroni-etal-2019-language, sung-etal-2021-language, bouraoui2019inducing}, we adopt a more comprehensive approach by predicting multi-token entities. Following \citet{kassner-etal-2021-multilingual}, we frame the task as entity ranking, but with a key difference: we do not limit predictions to specific entity types. This is because entity type information is often scarce or unreliable, and using external resources to classify entity types may introduce inaccuracies. Instead, following \citet{shaier-etal-2024-comparing}, we allow our models to predict any entity from the dataset's entity list.
To evaluate the predictions, we use the top-k accuracy metric, as suggested by \citet{sung-etal-2021-language, shaier-etal-2024-comparing}. This metric assigns a score of 1 if the correct entity is among the top k predicted entities, and 0 otherwise. Due to the complex relationships between entities (N-to-M connections), we report accuracy at three different levels: Acc@1, Acc@5, and Acc@10.

\textbf{CLM Evaluation:} 
Assessing the accuracy of CLM-generated answers is a complex task, especially since a proportion of responses are lengthy and explanatory. As a result, the traditional exact match metric commonly used in question-answering tasks is inadequate. We therefore adopt best sub-span accuracy, following \citet{kandpal2023large, mallen2023not}, which is equal to 1 if the gold label appears in the generated answer and 0 otherwise. To account for different capitalization possibilities in labels and outputs, we cast both to lowercase before scoring the outputs.

To mitigate the known sensitivity of CLMs to minor input variations \cite{jia-liang-2017-adversarial, shaier-etal-2024-say}, we employ three diverse prompt designs (detailed in Appendix \ref{CLM_prompts}), in line with established CLM probing research \cite{taylor2022galactica, li-etal-2023-language-models, Kalo_2022, Nayak_2023}. Notably, previous studies have shown that varying prompt designs, such as providing in-context examples, can significantly enhance performance \cite{NEURIPS2020_1457c0d6}. To this end, one of our prompts includes five in-context examples. To maintain consistency with the MLM input, we opt for a zero-shot evaluation, where no examples or additional context are provided with the other two prompts. 

Given the substantial resources required to evaluate the larger models (GPT-4 and Llama 405B), we first assess the three prompts using the smaller models on a representative sample of the data. Our results show that the in-context learning prompt performs best. We therefore evaluate the larger models on this prompt on the vast English data, and all models on the remaining languages data.

\section{Results \& Discussion}
Due to space constraints, we provide a summary of our results in Table \ref{all-lms}. See full presentation of models' results in Appendix \ref{appendix_tables}.

\subsection{Model-wise Analysis} 
For CLMs, Llama 3.1 405b Instruct performs best in all languages. On English sentence-level, it scores $57.5\%$, while GPT-4o-mini scores the lowest at $49.5\%$. On paragraph-level, Llama 3.1 405b Instruct scores $67.0\%$. Similar results are seen in Spanish (Llama 3.1: $49.5\%$, GPT-4o-mini: $37.4\%$) and Polish (Llama 3.1: $39.6\%$, GPT-4o-mini: $27.0\%$).

For MLM, XLM-RoBERTa excels on English sentence- and paragraph-level, with top-1, top-5, and top-10 scores of $23.8`\%$, $45.6\%$, and $54.2\%$, respectively. mBERT performs the worst, with scores of $12.5\%$, $33.4\%$, and $41.9\%$, respectively. XLM-RoBERTa dominates on Spanish and Polish sentence- and paragraph-level, with scores of $17.8\%$, $40.5\%$, and $49.9\%$ on Spanish, and similar results on Polish.

While these models show promise, their educational knowledge is limited, and their performance varies greatly between languages, particularly in CLMs. This language gap is problematic in educational settings, where students from diverse linguistic backgrounds need equal access to educational resources. Many students speak languages other than English, and the significant performance disparity between languages may exacerbate existing educational inequalities. 

For instance, students who speak Spanish or Polish may not receive the same level of educational support as their English-speaking peers, simply because the models are less proficient in their native languages. Notably, this gap is smaller in MLM, but it is unclear if multilingual models perform equally well across all subjects due to the larger English dataset. Ultimately, bridging this language gap is crucial to ensuring that AI-powered educational tools are inclusive, equitable, and effective for all students.

\begin{table*}
\centering
\setlength{\tabcolsep}{0.6pt}
\tiny
\resizebox{\textwidth}{!}{%
\begin{tabular}{l|*{11}{>{\centering\arraybackslash}m{1.5cm}}}
\hline
\multicolumn{1}{c|}{\textbf{Models}} & 
\multicolumn{6}{c}{\textbf{Languages}} \\
& \multicolumn{2}{c}{\textbf{English}} & \multicolumn{2}{c}{\textbf{Spanish}} & \multicolumn{2}{c}{\textbf{Polish}}\\

& \textbf{S} & \textbf{P} & \textbf{S} & \textbf{P} & \textbf{S} & \textbf{P}\\
\hline

GPT-4-turbo & 55.1 & 62.2 & 35.8 & 43.4 & 22.9 & 31.3 \\
GPT-4o-mini & 49.5 & 59.8 & 30.1 & 39.6 & 18.6 & 28.1 \\
GPT-4o & 53.4 & 63.5 & 34.5 & 42.7 & 27.7 & 38.3 \\
Llama-3.1-405B-Instruct & 57.5 & \textbf{67.0} & 42.4 & \textbf{51.4} & 27.8 & \textbf{39.7} \\
\hline

BERT & [15.4, 32.6, 40.4] & [16.7, 34.6, 41.1] & - & - & - & -\\
DistillBERT & [13.0, 32.3, 41.8] & [14.2, 35.1, 43.8] & - & - & - & - \\
XLM-RoBERTa & [18.7, 40.0, 50.3] & [\textbf{23.8}, 45.6, \textbf{54.2}] & [12.5, 31.1, 42.7] & [\textbf{17.8}, \textbf{40.5}, \textbf{49.9}] & [11.2, 29.4, 38.1] & [\textbf{29.9}, \textbf{51.7}, \textbf{59.4}]\\
mBERT & [9.3, 27.4, 36.8] & [12.5, 33.4, 41.9] & [9.1, 28.4, 38.1] & [13.3, 36.3, 45.5] & [4.4, 17.4, 24.2] & [16.1, 37.3, 45.7]\\
SciBERT & [18.5, 40.9, 49.6] & [22.6, \textbf{45.7}, 53.9] & - & - & - & - \\

\hline
\end{tabular}%
}
\caption{\label{all-lms} All models' overall accuracies (percentages) \textbf{on the English, Spanish, and, Polish portions of the dataset}. Results are shown as [prompt 1, prompt 2, prompt 3] for Causal LMs, and [top1, top5, top10] for Masked LMs. S=sentence-level prompts; P=paragraph-level prompts. The highest result for each setting, language, and model type is highlighted in \textbf{bold} font.}
\end{table*}

\subsection{Prompt Design and Evaluation} 
One striking observation is that the performance of most CLMs varies significantly depending on the prompt used for evaluation. For instance, on the Spanish portion, Llama 3's performance on Business Statistics diverges drastically across different prompt settings (see Table \ref{causal-lms-spanish}): it achieves a score of $23.0\%$ when evaluated using paragraph-level prompts with setting 3, but only $5.0\%$ when evaluated with prompt 1. This notable discrepancy underscores the importance of evaluating CLMs using a diverse range of prompts to gain a comprehensive understanding of their knowledge. 

\subsection{Sentence-Level vs. Paragraph-Level Performance}  
As discussed in \citet{petroni-etal-2019-language}, CS probing evaluates the lower bound of LMs' knowledge. Notably, \citet{shaier-etal-2024-comparing} find that using template-free prompts instead of template-based prompts can reveal a higher lower bound of knowledge, as templates can bias model predictions.

Building upon this insight, MALAMUTE introduces a significant innovation by providing a comprehensive set of both sentence-level and paragraph-level template-free prompts. This enables a more nuanced evaluation of LMs' knowledge in diverse contextual settings. Our experimental findings demonstrate that, across nearly all subdomains and models, paragraph-level prompts consistently yield higher scores for both CLMs and MLMs. This suggests that the provision of additional context can substantially enhance the evaluation of knowledge and masked concepts.

Furthermore, MALAMUTE's extensive collection of parallel sentence and paragraph-level prompts allows us to draw a more definitive conclusion: LMs may possess a significantly higher level of knowledge than previously estimated. Notably, when the same concept is being masked, our results show that models exhibit a substantially higher likelihood of correct prediction when more context is provided. This finding has important implications for the development of more accurate and comprehensive knowledge evaluation frameworks for LMs.


\subsection{Granularity Analysis} 
The MALAMUTE benchmark represents a significant milestone in probing datasets, offering an unprecedented level of granularity that enables a fine-grained examination of LMs' strengths and weaknesses. This level of detail is crucial for developing effective educational systems that interact with students across diverse levels and domains.

For instance, our analysis of CLMs reveals that the top-performing model, Llama 3.1 405b Instruct, excels in subjects like Concepts of Biology, American Government, and Anatomy with scores approaching $80\%$ (see Table \ref{causal-lms-english}). However, it struggles with subjects like Calculus 3, Marketing, and Business Law, where its scores plummet to around $50\%$. Similarly, MLMs demonstrate varying degrees of proficiency across subjects, with some showing significant knowledge gaps.

The granular nature of MALAMUTE is essential because it allows us to uncover hidden patterns and biases in model performance. While models may appear successful at first glance, a closer inspection can reveal significant gaps in their knowledge. For example, SciBERT, the overall top-performing MLM, owes its success largely to its exceptional performance in scientific subjects, with average scores of around $60\%$ in Math and Science. However, when we drill down to specific subjects like Philosophy, World History, or Anthropology, its scores drop to around $30\%$. Furthermore, despite its strong performance in certain science domains, SciBERT's knowledge in Astronomy is found to be lacking.

The insights provided by MALAMUTE underscore the critical need for more targeted and fine-grained evaluation and development of LMs. By recognizing the specific areas where models excel and struggle, we can focus our efforts on creating more effective, well-rounded models that can support students across diverse educational domains.

\subsection{MLM vs. CLM Performance}
Notably, many MLMs outperform many CLMs across all languages and subjects, which is a surprising finding given the significant size difference between the two types of models. Specifically, the smallest CLM (with 7B parameters) is more than 10 times larger than the largest MLM (XLM-RoBERTa). However, it is essential to acknowledge that this comparison is not apples-to-apples, as the two types of models are evaluated differently. MLMs are evaluated on their ability to predict masked tokens, which has a relatively limited number of possible options, whereas CLMs are tasked with generating free-form text, which has an exponentially larger solution space.

Furthermore, the entity ranking evaluation method used for MLMs is simpler and more straightforward compared to the sub-span accuracy method used for CLMs. This raises important questions: (1) Can we directly compare the knowledge captured by MLMs and CLMs, given their fundamentally different evaluation criteria? and (2) What alternative evaluation methods, yet to be explored, could provide a more accurate assessment of each model type's knowledge? We present these findings and questions for the reader's consideration, acknowledging the complexity and nuance of comparing these two types of LMs.

\section{Conclusion}

We introduce MALAMUTE, a highly-granular, multilingual, and template-free educational probing dataset that exposes significant knowledge gaps in existing LMs. As LMs are increasingly considered for classroom use, where their performance can directly impact student outcomes, it is essential to develop evaluation methods that assess knowledge in specific, curriculum-aligned areas. MALAMUTE highlights the need for more specialized benchmarks that go beyond general knowledge and ensure models can support, rather than hinder, student learning. We hope MALAMUTE inspires the development of more educational benchmarks, driving the field toward safer, more reliable LMs for educational contexts. By addressing these gaps, we can better align language models with the needs of educational systems and improve their utility in real-world applications.

\section*{Limitations} 

We acknowledge that student materials are subject to change over time. However, we have mitigated this issue by using OpenStax, a constantly updated and expanding library of open educational resources. With new subjects, textbook improvements, and language translations being added regularly, our dataset will remain adaptable and extensible in the future.

\section*{Ethics Statement}
We developed MALAMUTE to promote the creation of safer educational systems. We believe it is essential to rigorously evaluate language models on real-world student materials before deploying them in actual educational settings, to ensure they can accurately and reliably support student learning.

\bibliography{anthology,custom}

\clearpage

\appendix

\section{Causal LMs Prompts}
\label{CLM_prompts}

We evaluate the performance of our CLMs on the MALAMUTE dataset using each of the following three prompts. We represent each masked prompt as $p_{ti}$.

\subsection{Prompt 1}
The following prompt is based on \citet{taylor2022galactica,li-etal-2023-language-models}'s work

\noindent

\begin{center}
\texttt{$p_{ti}$}
\end{center}

\subsection{Prompt 2}
The following prompt is based on \citet{luo-etal-2023-systematic}'s work

\begin{center}
\texttt{
Please predict the missing words to complete the following sentence based on the facts in the real-world. The missing words are represented by [MASK]. Please return the missing words only. $p_{ti}$ }
\end{center}

\subsection{Prompt 3}
We drew inspiration from the work of \citet{Nayak_2023} and created a modified prompt. Specifically, we augmented the original prompt by incorporating five in-context examples, which are designed to provide additional context and guidance.

\begin{center}
\texttt{
You are a helpful, respectful and honest assistant. Your answers should be crisp, short and not repetitive.\
Here are 5 examples: Example 1: Prompt: Francesco Bartolomeo Conti was born in [MASK]. Answer: Florence. Example 2: Prompt: English bulldog is a subclass of [MASK]. Answer: dog. Example 3: Prompt: The official language of Mauritius is [MASK]. Answer: English. Example 4: Prompt: Nicotine binds to [MASK]. Answer: CHRNA4. Example 5: Prompt: Hepatitis has symptoms such as [MASK]. Answer: Abdominal pain. Prompt: $p_{ti}$}
\end{center}

\section{Filtration Methodology}
\label{filtration}

To ensure the quality of our prompts, we have developed a comprehensive filtration methodology that involves a combination of regular expressions and POS tagging. This methodology enables us to remove prompts that do not meet our quality standards, as determined through a thorough manual review of the prompt set.

Our filtration methodology begins with the application of regular expressions to remove prompts that exhibit commonly seen issues. Specifically, we remove prompts that start with the following words: "this'', ``that'', ``and'', ``those'', ``these'', ``such'', ``we'', ``also'', ``it'', ``for example'', ``even'', ``they'', ``here'', ``another'', ``hence'', ``therefore'', ``neither'', and "either". We also remove prompts that have digits as labels, as well as those that contain fewer than five words. Additionally, we eliminate prompts that refer to visual aids, including "graphs'', ``images'', ``figures'', ``plots'', ``example'', ``diagram'', ``chart'', ``photograph'', and "photo".

Furthermore, we remove prompts that start with or contain action words, including: "explain'', ``describe'', ``analyze'', ``discuss'', ``interpret'', ``summarize'', ``define'', ``identify'', ``clarify'', ``elucidate'', ``examine'', ``determine'', ``decide'', ``solve'', ``calculate'', ``evaluate'', ``assess'', ``judge'', ``compare'', ``contrast'', ``distinguish'', ``differentiate'', ``write'', ``create'', ``design'', ``develop'', ``formulate'', ``compose'', ``draw'', ``sketch'', ``graph'', ``plot'', ``take'', ``find'', ``state'', ``prove'', ``illustrate'', ``represent'', ``model'', ``simulate'', ``demonstrate'', ``show'', ``indicate'', ``point'', ``outline'', ``develop'', ``expand'', ``elaborate'', ``justify'', ``argue'', ``debate'', ``review'', ``revise'', ``edit'', ``redraft'', ``reorganize'', ``rephrase'', ``paraphrase'', ``translate'', ``transcribe'', ``record'', ``document'', ``report'', ``present'', ``display'', ``exhibit'', ``construct'', ``build'', ``make'', ``produce'', ``generate'', ``develop'', ``plan'', ``organize'', ``schedule'', ``coordinate'', ``arrange'', ``implement'', ``execute'', ``enact'', ``establish'', ``institute'', ``initiate'', ``begin'', ``start'', ``continue'', ``resume'', ``complete'', ``finish'', and "conclude".

We also use regular expressions to remove parenthesized text that appears immediately after or at the end of a label, such as "(AC)" in "audit committee (AC)". However, we apply this step with caution, ensuring that the parenthesized text is separated from the label by a white space character. This constraint is necessary to avoid damaging prompts in subjects like mathematics and computer science, where parentheses are frequently used to denote other concepts, such as mathematical notation (e.g., ``f(x)") or programming language syntax (e.g., ``upper()").

Our comprehensive filtration methodology enables us to remove low-quality prompts and ensure that our dataset meets the highest standards of quality and accuracy.

\section{Quality Control Methodology}
\label{quality_control}
To ensure the high quality of the generated prompts, we conducted a rigorous quality control process. This process involved manually reviewing a representative sample of 495 English prompts, consisting of 9 prompts from each of the 55 subdomains. The sample included a mix of sentence-level and paragraph-level prompts, with a ratio of 2:1, to provide a comprehensive representation of our dataset.

Three graduate student annotators, each holding an undergraduate degree and possessing in-depth knowledge of MALAMUTE's diverse subject areas, evaluated the prompts based on a standardized set of criteria. Their knowledge and expertise in these areas are evident from their academic transcripts, which demonstrate a strong foundation in the subject matter. This evaluation was designed to assess two key aspects of the prompts: (1) grammatical correctness, and (2) the effective description of concept properties, which enables the inference of the label from the context.

The results of our quality control process revealed an impressive level of quality in the generated prompts. Among the paragraph-level prompts, a substantial $97.6\%$ were found to be grammatically correct, while a significant $92.9\%$ effectively conveyed the intended concept. Similarly, the sentence-level prompts demonstrated a high level of quality, with $99.1\%$ meeting grammatical standards and $93.9\%$ successfully demonstrating the concept. It is important to note that in many probing datasets, entities are often related to numerous other entities (N-to-M connections), which can result in multiple possible answers \cite{sung-etal-2021-language, shaier-etal-2024-comparing}. Our results indicate that while there is some variation in quality, the overwhelming majority of prompts are suitable for knowledge probing applications.

It is worth noting that our approach to selecting annotators was deliberate. We chose to select annotators who were non-experts in any particular field of study, rather than domain-specific experts. This decision was motivated by two key considerations. Firstly, given the highly granular nature of MALAMUTE's knowledge, it would be impractical to find and engage a large number of annotators with expertise in each of the domains, subdomains, and concepts. Secondly, and more importantly, LLMs are trained on general data and are, in that sense, also general and "non-experts." Therefore, it is more suitable to evaluate the performance of models against humans who possess general knowledge, but also have a focus on MALAMUTE's highly granular knowledge in their studies.

By conducting this rigorous quality control process, we were able to verify that the generated prompts meet the desired standards of quality, and are suitable for knowledge probing applications. While there may be some variation in quality, the overwhelming majority of prompts demonstrate a high level of quality, providing a robust foundation for evaluating the performance of LLMs.

\subsection{Annotation Prompts}
For each prompt, annotators were instructed to evaluate two aspects in response to the following prompt: ``For each prompt evaluate whether: The prompt is grammatically correct; The label is specific to the prompt, i.e. the label can be inferred from the context and there aren't many possible alternative answers. Use web resources such as Google search, Wikipedia, etc., as needed.'' Annotators were encouraged to use their discretion and consider all relevant information when making their judgments. Importantly, there were no time constraints imposed on the annotation task, allowing annotators to take as much time as necessary to provide accurate assessments.


\section{Additional Results}
The complete results of our evaluation are presented in the following tables. Table \ref{causal-lms-english} shows the performance of CLMs on English data. Table \ref{masked-lms-eng-sentence} presents the results of MLMs evaluated on English at the sentence level, while Table \ref{masked-lms-eng-para} provides their performance at the paragraph level. For Spanish, Tables \ref{causal-lms-spanish} and \ref{more_causal-lms-spanish} summarize the results of CLMs, and Tables \ref{masked-lms-spanish} details the MLM evaluations. Similarly, for Polish, Tables \ref{causal-lms-polish} and \ref{more_causal-lms-polish} report the performance of CLMs, and Table \ref{masked-lms-polish} presents the results of MLMs.

\label{appendix_tables}
\begin{table*}
\centering
\setlength{\tabcolsep}{1.2pt}
\tiny
\begin{tabular}{l|*{9}{>{\centering\arraybackslash}m{1.5cm}}}
\hline

\multicolumn{1}{c|}{\textbf{Subdomain}} & 

\multicolumn{8}{c}{\textbf{Causal LMs}} \\
&  \multicolumn{2}{c}{\textbf{GPT-4-Turbo}} & \multicolumn{2}{c}{\textbf{GPT-4o-mini}} & \multicolumn{2}{c}{\textbf{GPT-4o}} & \multicolumn{2}{c}{\textbf{Llama 3.1 405b Instruct}} \\
&  \textbf{S} & \textbf{P} & \textbf{S} & \textbf{P} & \textbf{S} & \textbf{P} & \textbf{S} & \textbf{P} \\

\hline
        Prealgebra & 59.4 & 61.9 & 59.4 & 59.6 & 59.0 & 63.5 & 64.1 & \textbf{70.5} \\
        Elementary Algebra & 48.5 & 53.0 & 49.1 & 54.3 & 45.5 & 55.3 & 55.7 & \textbf{61.3} \\
        Intermediate Algebra & 49.1 & 55.4 & 46.0 & 54.4 & 50.3 & 52.9 & 53.4 & \textbf{60.2} \\
        College Algebra & 51.0 & 53.3 & 49.7 & 58.8 & 53.1 & 58.8 & 58.6 & \textbf{63.3}\\
        Trigonometry & 49.8 & 55.9 & 49.5 & 56.2 & 51.6 & 61.3 & 56.3 & \textbf{64.5} \\
        Precalculus $\textcolor{purple}\star$ & 46.4 & 49.9 & 44.9 & 49.8 & 46.6 & 52.2 & 54.9 & \textbf{59.2} \\
        Calculus I $\textcolor{purple}\star$ & 47.3 & 51.6 & 39.3 & 50.0 & 48.2 & 49.5 & 56.2 & \textbf{58.2} \\
        Calculus II $\textcolor{purple}\star$ & 46.2 & 51.0 & 41.9 & 45.2 & 45.6 & 50.2 & 48.1 & \textbf{53.3} \\
        Calculus III $\textcolor{purple}\star$ & 36.6 & 39.1 & 37.8 & 37.7 & 37.8 & 43.9 & 45.1 & \textbf{49.5} \\
        Intro. Statistics & 41.9 & 59.1 & 41.9 & 60.6 & 43.5 & 56.2 & 53.2 & \textbf{68.6} \\
        Business Statistics $\textcolor{purple}\star$ & 44.1 & 59.2 & 45.4 & 57.8 & 46.1 & 59.7 & 52.0 & \textbf{64.5} \\
        Statistics $\textcolor{purple}\star$ & 41.2 & 54.6 & 42.5 & 55.3 & 41.2 & 53.3 & 55.0 & \textbf{61.2} \\
        Contemporary Math & 51.8 & 62.0 & 50.0 & 62.3 & 49.6 & 63.0 & 56.7 & \textbf{65.0} \\
\hline
Math Avg.  & 47.2 & 54.3 & 46.0 & 54.0 & 47.5 & 55.4 & 54.6 & \textbf{61.5} \\

\hline
\hline
        Concepts of Biology & 71.2 & 79.9 & 67.0 & 78.7 & 72.7 & 82.4 & 73.9 & \textbf{83.7} \\
        Biology & 66.0 & 74.2 & 61.3 & 74.1 & 68.3 & 76.6 & 68.1 & \textbf{78.2} \\
        Microbiology & 56.0 & 62.6 & 50.4 & 60.4 & 56.0 & 64.8 & 58.1 & 68.1 \\
        Chemistry Atoms First & 64.4 & 68.7 & 61.0 & 69.3 & 66.0 & 72.9 & 68.8 & \textbf{77.2} \\
        Chemistry $\textcolor{purple}\star$ & 64.3 & 73.3 & 50.0 & 73.3 & 64.3 & 75.6 & 67.9 & \textbf{77.8} \\
        Organic Chemistry & 61.2 & 64.9 & 56.9 & 65.3 & 65.4 & 69.6 & 65.6 & \textbf{72.0} \\
        Anatomy & 62.4 & 72.5 & 58.0 & 72.9 & 66.1 & 77.8 & 65.7 & \textbf{79.0} \\
        Astronomy & 66.3 & 73.5 & 57.6 & 67.5 & 63.9 & 72.5 & 66.9 & \textbf{75.3} \\
        College Physics & 57.2 & 62.6 & 51.5 & 62.6 & 55.6 & 63.5 & 59.0 & \textbf{66.8} \\
        Physics & 62.5 & 67.6 & 60.3 & 68.3 & 61.9 & 69.2 & 66.2 & \textbf{72.8} \\
        Physics I $\textcolor{red}\diamond$ $\textcolor{purple}\star$ & 40.1 & 49.2 & 37.2 & 47.0 & 38.9 & 49.1 & 43.0 & \textbf{53.8} \\
        Physics II $\textcolor{red}\diamond$ $\textcolor{purple}\star$ & 49.9 & 58.4 & 43.5 & 56.7 & 50.6 & 56.4 & 54.2 & \textbf{61.6} \\
        Physics III $\textcolor{red}\diamond$ $\textcolor{purple}\star$ & 48.4 & 53.3 & 43.9 & 52.0 & 49.5 & 55.6 & 57.3 & \textbf{60.5} \\
\hline
Science Avg.  & 59.2 & 66.2 & 53.7 & 65.2 & 59.9 & 68.2 & 62.7 & \textbf{71.3} \\

\hline
\hline
        U.S. History & 65.8 & 74.4 & 57.0 & 72.2 & 64.8 & 74.8 & 68.3 & \textbf{77.0} \\
        Writing & 46.8 & 56.6 & 41.1 & 51.3 & 49.0 & 58.6 & 48.4 & \textbf{58.5} \\
        Philosophy & 59.6 & 68.3 & 50.8 & 64.3 & 57.4 & 69.3 & 59.9 & \textbf{72.5} \\
        World History I & 58.7 & 64.6 & 49.2 & 60.7 & 57.8 & 64.7 & 59.5 & \textbf{67.8} \\
        World History II & 65.7 & 70.2 & 58.5 & 67.2 & 64.7 & 70.3 & 66.6 & \textbf{72.9} \\
\hline
Humanities Avg.  & 59.3 & 66.8 & 51.3 & 63.1 & 58.7 & 67.5 & 60.5 & \textbf{69.7} \\

\hline
\hline
        Intro. to Business & 43.8 & 52.2 & 36.2 & 48.8 & 42.7 & 52.2 & 45.1 & \textbf{56.2} \\
        Financial Accounting & 47.4 & 54.0 & 44.7 & 57.5 & 48.4 & 56.3 & 50.0 & \textbf{62.8} \\
        Managarial Accounting & 40.9 & 50.1 & 37.4 & 50.4 & 42.9 & 52.2 & 43.1 & \textbf{59.0} \\
        Business Ethics & 51.9 & 63.1 & 45.0 & 57.5 & 50.4 & 57.1 & 54.7 & \textbf{66.4} \\
        Org. Behavior & 31.2 & 47.8 & 27.7 & 42.5 & 30.7 & 47.8 & 34.7 & \textbf{51.6} \\
        Finance & 44.3 & 51.0 & 39.9 & 49.5 & 43.8 & 50.7 & 48.5 & \textbf{58.8} \\
        Business Law & 41.7 & 51.3 & 41.7 & 52.5 & 43.3 & 52.8 & 46.8 & \textbf{57.2} \\
        Intellectual Property & 47.2 & 53.8 & 42.7 & 48.7 & 44.9 & 52.1 & 44.9 & \textbf{55.5} \\
        Marketing & 38.9 & 46.8 & 33.7 & 43.4 & 38.5 & 47.2 & 39.8 & \textbf{48.5} \\
        Management & 41.4 & 53.4 & 40.1 & 53.0 & 41.4 & 53.0 & 41.9 & \textbf{57.3} \\
        Economics & 46.8 & 58.2 & 45.0 & 57.6 & 47.7 & 58.8 & 53.0 & \textbf{66.8} \\
        Macroeconomics $\textcolor{red}\diamond$ & 56.9 & 63.3 & 54.3 & 64.0 & 57.8 & 64.7 & 61.2 & \textbf{73.4} \\
        Microeconomics $\textcolor{red}\diamond$ & 59.7 & 69.6 & 61.3 & 73.4 & 62.9 & 70.9 & 71.0 & \textbf{78.5} \\
        Entrepreneurship & 46.4 & 54.5 & 39.0 & 48.3 & 46.7 & 54.2 & 48.4 & \textbf{57.9} \\
\hline
Business Avg.  & 45.6 & 54.9 & 42.1 & 53.4 & 45.9 & 55.0 & 48.8 & \textbf{60.7} \\

\hline
\hline
        Nutrition & 43.0 & 48.0 & 39.0 & 46.4 & 42.1 & 48.3 & 43.7 & \textbf{52.8} \\
\hline
Nursing Avg.  & 43.0 & 48.0 & 39.0 & 46.4 & 42.1 & 48.3 & 43.7 & \textbf{52.8} \\

\hline
\hline
        Python Programming &  46.0 & 55.2 & 46.8 & 56.5 & 51.0 & 55.2 & 50.1 & \textbf{61.8} \\
        Workplace Software & 37.4 & 52.9 & 34.4 & 49.9 & 38.1 & 54.6 & 39.6 & \textbf{59.7} \\
\hline
Computer Science Avg.  & 41.7 & 54.0 & 40.6 & 53.2 & 44.6 & 54.9 & 44.8 & \textbf{60.8} \\

\hline
\hline
        American Government & 66.0 & 76.1 & 61.0 & 72.7 & 64.9 & 75.8 & 70.6 & \textbf{80.2} \\
        Anthropology & 55.4 & 59.7 & 44.3 & 52.1 & 55.9 & 61.0 & 57.8 & \textbf{65.5} \\
        Political Science & 58.6 & 64.6 & 52.0 & 59.6 & 57.5 & 64.9 & 59.8 & \textbf{67.7} \\
        Psychology $\textcolor{red}\diamond$ & 66.4 & 73.7 & 62.9 & 74.6 & 68.7 & \textbf{78.1} & 67.2 & 78.0 \\
        Sociology & 58.4 & 68.9 & 49.2 & 66.1 & 60.1 & 71.3 & 59.9 & \textbf{73.3} \\
\hline 
Social Sciences Avg.  & 61.0 & 68.6 & 53.9 & 65.0 & 61.4 & 70.2 & 63.1 & \textbf{72.9} \\

\hline
\hline
        College Success & 46.5 & 60.9 & 37.2 & \textbf{65.6} & 44.2 & 62.5 & 41.9 & 57.8 \\
        Col. Suc. Concise & 40.4 & 53.8 & 40.4 & 61.5 & 42.6 & 51.9 & 44.7 & \textbf{67.3} \\
\hline

College Success Avg.  & 43.4 & 57.4 & 38.8 & \textbf{63.6} & 43.4 & 57.2 & 43.3 & 62.6 \\

\hline
\hline
Overall & 51.6 & 59.7 & 47.4 & 58.5 & 52.0 & 60.7 & 55.4 & \textbf{65.3} \\
\hline
\hline

\end{tabular}
\caption{\label{causal-lms-english} Causal models' accuracies (percentages) \textbf{on the English portion of the dataset}. Books that are also in Polish are marked with $\textcolor{red}\diamond$, where books that are also in Spanish are marked with $\textcolor{purple}\star$, and their results can be seen in Tables T1, T2. S=sentence-level prompts; P=paragraph-level prompts. The highest result for each subdomain is highlighted in \textbf{bold} font.}
\end{table*}

\begin{table*}
\centering
\setlength{\tabcolsep}{0.6pt}
\tiny
\begin{tabular}{l|*{11}{>{\centering\arraybackslash}m{1.5cm}}}
\hline
\multicolumn{1}{c|}{\textbf{Subdomain}} & 
\multicolumn{5}{c}{\textbf{Mask LMs}} \\
& \multicolumn{1}{c}{\textbf{BERT}} & \multicolumn{1}{c}{\textbf{DistilBERT}} & \multicolumn{1}{c}{\textbf{XLM-RoBERTa}} & \multicolumn{1}{c}{\textbf{mBERT}} & \multicolumn{1}{c}{\textbf{SciBERT}} \\

\hline
Prealgebra & [15.2, 37.8, 46.5] & [14.7, 36.4, 47.9] & [19.4, 37.8, 47.9] & [8.8, 25.8, 36.9] & [\textbf{23.5}, \textbf{48.4}, \textbf{57.6}] \\  
Elementary Algebra & [15.0, 39.5, 49.7] & [13.8, 38.9, 49.7] & [18.6, 40.7, 53.3] & [6.0, 25.1, 38.3] & [\textbf{22.8}, \textbf{52.1}, \textbf{60.5}] \\
Intermediate Algebra & [10.4, 27.6, 37.4] & [12.3, 31.3, 42.9] & [13.5, 30.7, 39.3] & [6.1, 28.8, 36.2] & [14.7, 41.7, 50.3] \\
College Algebra & [9.7, 33.1, 42.1] & [9.7, 36.6, 47.6] & [7.6, 31.7, 42.8] & [4.1, 23.4, 30.3] & [20.0, \textbf{53.8}, \textbf{63.4}] \\
Trigonometry & [10.1, 28.5, 34.3] & [9.7, 29.6, 36.8] & [11.2, 27.8, 37.9] & [4.3, 18.4, 27.4] & [\textbf{20.2}, 41.5, 52.0] \\
Precalculus $\textcolor{purple}\star$ & [8.3, 27.6, 33.0] & [9.2, 29.2, 37.2] & [9.0, 29.4, 38.4] & [3.8, 21.5, 31.3] & [14.8, 42.6, 51.6] \\
Calculus I $\textcolor{purple}\star$ & [19.6, 34.8, 46.4] & [17.0, 44.6, 53.6] & [14.3, 33.0, 50.9] & [13.4, 30.4, 45.5] & [23.2, 55.4, \textbf{65.2}] \\
Calculus II $\textcolor{purple}\star$ & [10.6, 30.0, 36.2] & [11.9, 35.0, 43.8] & [13.8, 35.0, 44.4] & [10.0, 28.7, 39.4] & [23.8, 45.0, 54.4] \\
Calculus III $\textcolor{purple}\star$ & [13.4, 36.6, 47.0] & [14.0, 38.4, 45.7] & [12.8, 36.6, 47.6] & [9.1, 28.7, 37.2] & [\textbf{23.2}, \textbf{50.0}, \textbf{61.0}] \\  
Intro. Statistics & [16.1, 40.3, 50.0] & [16.1, 37.1, 61.3] & [19.4, 40.3, 59.7] & [14.5, 37.1, 53.2] & [\textbf{40.3}, \textbf{75.8}, \textbf{83.9}] \\
Business Statistics $\textcolor{purple}\star$ & [12.5, 36.2, 48.0] & [11.8, 36.8, 48.0] & [13.2, 34.2, 46.7] & [5.9, 28.3, 38.2] & [27.0, 55.9, 66.4] \\
Statistics $\textcolor{purple}\star$ & [16.2, 41.2, 51.2] & [16.2, 43.8, 55.0] & [16.2, 43.8, 61.3] & [10.0, 38.8, 48.8] & [28.7, \textbf{70.0}, \textbf{80.0}] \\
Contemporary Math & [18.7, 44.7, 56.0] & [21.1, 49.3, 62.0] & [18.3, 47.5, 62.0] & [9.9, 35.9, 45.4] & [25.4, 54.6, 66.9] \\

\hline

Math Avg. & [13.5, 35.2, 44.4] & [13.7, 37.5, 48.6] & [14.4, 36.0, 48.6] & [8.1, 28.5, 39.1] & [23.7, 52.8, 62.6] \\

\hline
\hline
Concepts of Biology & [11.4, 22.3, 26.6] & [10.9, 28.1, 37.9] & [14.5, 33.0, 41.3] & [6.7, 24.6, 32.1] & [26.9, 53.4, 64.8] \\
Biology & [7.8, 16.9, 20.7] & [7.4, 19.8, 26.1] & [10.5, 24.7, 31.2] & [4.7, 18.0, 23.2] & [19.5, 42.6, 50.7] \\
Microbiology & [3.8, 8.4, 11.0] & [3.3, 10.1, 14.2] & [5.8, 16.5, 22.7] & [2.3, 11.8, 17.5] & [14.4, 32.8, 41.7] \\
Chemistry Atoms First & [14.1, 31.6, 39.0] & [10.7, 32.4, 41.1] & [14.5, 38.9, 49.5] & [9.2, 27.5, 38.1] & [20.3, 48.1, 56.9] \\
Chemistry $\textcolor{purple}\star$ & [32.1, 64.3, 67.9] & [25.0, 57.1, 82.1] & [32.1, \textbf{78.6}, \textbf{96.4}] & [21.4, 64.3, 92.9] & [35.7, 71.4, 89.3] \\
Organic Chemistry & [12.7, 22.8, 26.8] & [10.5, 28.6, 36.8] & [12.9, 36.2, 42.9] & [8.5, 25.9, 33.7] & [22.1, 49.6, 57.4] \\
Anatomy & [4.7, 10.4, 13.5] & [3.9, 12.3, 17.4] & [7.8, 20.8, 27.2] & [3.0, 16.1, 22.4] & [15.4, 38.0, 47.1] \\
Astronomy & [12.0, 27.4, 33.1] & [7.2, 22.4, 29.2] & [10.7, 27.7, 36.6] & [10.5, 25.3, 32.8] & [10.3, 23.0, 30.3] \\
College Physics & [11.4, 26.5, 33.5] & [10.7, 30.1, 39.3] & [12.8, 33.1, 41.1] & [6.3, 23.8, 31.7] & [19.9, 46.6, 57.0] \\
Physics & [18.3, 35.8, 46.6] & [15.1, 35.4, 43.7] & [19.2, 37.9, 49.4] & [9.1, 32.5, 40.9] & [23.9, \textbf{53.4}, 61.6] \\
Physics I $\textcolor{red}\diamond$ $\textcolor{purple}\star$ & [13.3, 27.8, 34.8] & [10.6, 29.2, 37.2] & [14.0, 32.4, 41.8] & [9.2, 23.7, 30.7] & [16.2, 41.8, \textbf{50.7}] \\
Physics II $\textcolor{red}\diamond$ $\textcolor{purple}\star$ & [13.7, 26.3, 34.9] & [12.4, 29.6, 35.4] & [18.7, 35.2, 45.6] & [10.4, 28.4, 35.9] & [21.3, 45.3, 51.9] \\
Physics III $\textcolor{red}\diamond$ $\textcolor{purple}\star$ & [12.9, 30.5, 35.7] & [13.6, 32.4, 40.8] & [15.7, 36.9, 44.6] & [10.8, 30.0, 39.9] & [26.1, 47.4, 54.5] \\

\hline
Science Avg. & [12.9, 27.0, 32.6] & [10.9, 28.3, 37.0] & [14.6, 34.8, 43.9] & [8.6, 27.1, 36.3] & [20.9, 45.6, 54.9]
\\

\hline
\hline

U.S. History & [16.3, 29.2, 34.6] & [15.0, 29.4, 35.2] & [18.3, 32.1, 39.5] & [14.0, 25.9, 32.5] & [8.4, 15.2, 21.6] \\
Writing & [15.2, 29.4, 36.4] & [11.8, 28.5, 38.2] & [20.9, 38.2, 48.0] & [9.1, 22.8, 29.8] & [10.6, 23.4, 30.2] \\
Philosophy & [17.6, 27.9, 33.1] & [11.7, 24.4, 31.2] & [16.3, 31.0, 39.0] & [13.1, 28.1, 35.1] & [9.8, 22.5, 28.7] \\
World History 1 & [13.1, 24.4, 28.9] & [7.6, 18.7, 24.9] & [9.0, 21.0, 27.0] & [7.8, 22.1, 30.4] & [3.0, 8.8, 12.7] \\
World History 2 & [23.6, 40.3, 47.9] & [10.5, 26.3, 34.7] & [20.9, 38.0, 46.3] & [15.9, 35.9, 45.2] & [8.8, 19.2, 24.9] \\

\hline

Humanities Avg. & [17.2, 30.2, 36.2] & [11.3, 25.5, 32.8] & [17.1, 32.1, 40.0] & [12.0, 27.0, 34.6] & [8.1, 17.8, 23.6] \\

\hline
\hline

Intro. to Business & [12.6, 23.6, 28.7] & [10.1, 21.6, 26.4] & [19.6, 36.4, 44.0] & [5.5, 15.8, 20.9] & [9.0, 21.7, 27.0] \\
Financial Accounting & [12.4, 32.1, 38.6] & [10.0, 35.0, 44.9] & [19.3, 47.4, 58.1] & [6.7, 22.2, 31.3] & [12.2, 31.5, 40.0] \\
Managarial Accounting & [14.6, 36.3, 42.9] & [11.4, 37.1, 48.3] & [22.6, 50.6, 63.1] & [9.1, 27.4, 36.9] & [16.0, 38.0, 45.1] \\
Business Ethics & [21.3, 41.9, 51.9] & [19.8, 42.6, 54.7] & [31.4, 56.2, 66.7] & [14.7, 31.8, 43.0] & [21.3, 45.7, 53.1] \\
Org. Behavior & [10.2, 25.9, 32.9] & [9.5, 24.9, 35.4] & [16.7, 38.4, 49.1] & [6.2, 19.2, 23.9] & [14.0, 33.2, 42.9] \\
Finance & [11.9, 27.4, 34.6] & [9.8, 25.3, 33.5] & [21.1, 44.0, 52.8] & [6.6, 22.1, 29.8] & [13.7, 30.6, 37.7] \\
Business Law & [27.0, 39.3, 44.8] & [21.8, 40.5, 46.8] & [29.0, \textbf{54.4}, 63.1] & [13.1, 31.3, 43.7] & [18.7, 38.9, 46.8] \\
Intellectual Property & [14.6, 46.1, 64.0] & [12.4, 38.2, 59.6] & [21.3, 49.4, \textbf{66.3}] & [7.9, 32.6, 44.9] & [9.0, 38.2, 57.3] \\
Marketing & [9.5, 20.5, 25.2] & [7.9, 20.2, 26.5] & [20.2, 38.3, 44.9] & [5.4, 15.5, 21.3] & [9.3, 20.5, 25.9] \\
Management & [21.2, 39.6, 52.3] & [14.9, 36.5, 46.8] & [28.8, 52.3, 64.4] & [14.9, 38.7, 47.3] & [21.2, 43.7, 53.6] \\
Economics & [11.9, 33.1, 41.6] & [10.4, 32.3, 42.9] & [17.2, 40.9, 50.5] & [5.2, 23.9, 31.6] & [15.8, 38.6, 48.8] \\
Macroeconomics $\textcolor{red}\diamond$ & [24.1, 58.6, 74.1] & [18.1, 56.0, 64.7] & [26.7, 59.5, 69.8] & [7.8, 43.1, 55.2] & [29.3, 60.3, 70.7] \\
Microeconomics $\textcolor{red}\diamond$ & [27.4, 64.5, 83.9] & [24.2, 61.3, 75.8] & [30.6, 66.1, 82.3] & [12.9, 53.2, 64.5] & [38.7, 71.0, 85.5] \\
Entrepreneurship & [10.0, 24.3, 30.0] & [8.4, 21.5, 28.5] & [18.0, 38.5, 45.2] & [5.8, 17.3, 22.9] & [8.6, 21.5, 28.3] \\

\hline

Business Avg. & [16.3, 36.7, 46.1] & [13.5, 35.2, 45.3] & [23.0, 48.0, 58.6] & [8.7, 28.2, 36.9] & [16.9, 38.1, 47.3] \\ 

\hline
\hline

Nutrition & [6.8, 15.7, 20.0] & [6.7, 16.3, 21.6] & [10.7, 27.5, 35.9] & [2.3, 10.6, 16.7] & [18.6, 39.8, 51.3] \\

\hline
Nursing Avg. & [6.8, 15.7, 20.0] & [6.7, 16.3, 21.6] & [10.7, 27.5, 35.9] & [2.3, 10.6, 16.7] & [18.6, 39.8, 51.3] \\

\hline
\hline
Python Programming & [7.6, 15.9, 20.1] & [6.8, 17.6, 20.7] & [20.1, 36.5, 46.2] & [4.0, 15.3, 23.5] & [16.4, 30.9, 34.6] \\
Workplace Software & [8.1, 17.2, 21.6] & [6.5, 16.7, 22.0] & [17.6, 36.3, 45.1] & [4.2, 13.0, 18.5] & [9.4, 22.8, 29.3] \\

\hline

Computer Science Avg. & [7.8, 16.6, 20.9] & [6.7, 17.1, 21.4] & [18.9, 36.4, 45.7] & [4.1, 14.2, 21.0] & [12.9, 26.9, 32.0] \\  

\hline
\hline

American Government & [22.8, 33.1, 38.3] & [10.4, 23.7, 31.0] & [23.4, 38.4, 45.5] & [15.8, 28.0, 33.4] & [7.0, 17.0, 22.3] \\
Anthropology & [12.3, 23.0, 29.0] & [9.5, 21.2, 28.4] & [13.0, 26.0, 32.9] & [8.6, 21.6, 28.0] & [8.2, 19.1, 24.9] \\
Political Science & [18.7, 32.8, 39.6] & [11.6, 26.4, 33.7] & [23.3, 41.6, 49.0] & [13.1, 26.5, 34.7] & [9.0, 22.3, 28.2] \\
Psychology $\textcolor{red}\diamond$ & [12.9, 26.4, 32.0] & [11.3, 26.8, 33.6] & [18.9, 39.0, 48.0] & [6.7, 20.6, 28.7] & [22.2, 42.6, 51.0] \\
Sociology & [19.1, 39.3, 48.9] & [16.8, 41.8, 54.0] & [27.9, 52.5, 63.9] & [9.9, 28.6, 38.4] & [17.2, 39.5, 49.8] \\

\hline
Social Sciences Avg. & [17.2, 30.9, 37.6] & [11.9, 28.0, 36.1] & [21.3, 39.5, 47.9] & [10.8, 25.1, 32.6] & [12.7, 28.1, 35.2] \\

\hline
\hline

College Success & [41.9, 60.5, 74.4] & [44.2, 62.8, 74.4] & [41.9, 76.7, 88.4] & [27.9, 53.5, 69.8] & [23.3, 55.8, 67.4] \\
College Success Concise & [40.4, 57.4, 70.2] & [29.8, 51.1, 70.2] & [46.8, 72.3, 87.2] & [17.0, 38.3, 59.6] & [27.7, 55.3, 63.8] \\

\hline
College Success Avg. & [41.1, 59.0, 72.3] & [37.0, 57.0, 72.3] & [44.3, 74.5, 87.8] & [22.4, 45.9, 64.7] & [25.5, 55.5, 65.6] \\

\hline
\hline
Overall & [15.4, 32.6, 40.4] & [13.0, 32.3, 41.8]  & [18.7, 40.0, 50.3] & [9.3, 27.4, 36.8] & [18.5, 40.9, 49.6] \\
\hline
\end{tabular}
\caption{\label{masked-lms-eng-sentence} Masked models' accuracies (percentages) \textbf{on the English portion of the dataset using sentence-level prompts}. Results are shown as [top1, top5, top10]. Books that are also in Polish are marked with $\textcolor{red}\diamond$, where books that are also in Spanish are marked with $\textcolor{purple}\star$. The highest result for each subdomain across both sentence and paragraph level Tables (Table \ref{masked-lms-eng-sentence} and Table \ref{masked-lms-eng-para}) is highlighted in \textbf{bold} font.}
\end{table*}
\begin{table*}
\centering
\setlength{\tabcolsep}{0.6pt}
\tiny
\begin{tabular}{l|*{11}{>{\centering\arraybackslash}m{1.5cm}}}
\hline
\multicolumn{1}{c|}{\textbf{Subdomain}} & 
\multicolumn{5}{c}{\textbf{Mask LMs}} \\
& \multicolumn{1}{c}{\textbf{BERT}} & \multicolumn{1}{c}{\textbf{DistilBERT}} & \multicolumn{1}{c}{\textbf{XLM-RoBERTa}} & \multicolumn{1}{c}{\textbf{mBERT}} & \multicolumn{1}{c}{\textbf{SciBERT}} \\

\hline
Prealgebra & [13.7, 35.0, 45.6] & [13.0, 33.7, 44.8] & [15.0, 37.0, 46.4] & [5.7, 26.9, 37.6] & [22.3, 47.2, 54.9] \\
Elementary Algebra & [12.8, 33.2, 41.9] & [13.1, 33.2, 41.9] & [15.7, 36.1, 45.7] & [8.6, 24.9, 33.9] & [22.4, 45.0, 55.9] \\
Intermediate Algebra & [11.6, 28.4, 38.2] & [9.8, 32.1, 44.3] & [11.3, 29.4, 39.1] & [7.6, 26.3, 34.6] & [\textbf{18.7}, \textbf{44.0}, \textbf{55.0}] \\
College Algebra & [11.6, 35.2, 44.2] & [13.1, 38.7, 47.7] & [11.6, 28.1, 42.2] & [4.5, 23.1, 30.7] & [\textbf{23.6}, 50.3, 59.8] \\
Trigonometry & [8.9, 29.6, 37.4] & [7.6, 32.0, 41.1] & [9.9, 27.3, 36.5] & [4.2, 22.2, 28.1] & [17.7, \textbf{42.1}, \textbf{53.2}] \\
Precalculus $\textcolor{purple}\star$ & [8.7, 26.5, 32.3] & [8.1, 28.4, 36.9] & [11.0, 28.2, 37.4] & [5.4, 22.8, 29.5] & [\textbf{17.5}, \textbf{43.8}, \textbf{52.6}] \\   
Calculus I $\textcolor{purple}\star$ & [13.6, 31.0, 40.8] & [14.1, 39.7, 47.8] & [18.5, 41.3, 48.4] & [11.4, 33.2, 39.7] & [\textbf{27.2}, \textbf{56.5}, 62.5] \\
Calculus II $\textcolor{purple}\star$ & [14.2, 33.7, 39.5] & [13.4, 36.0, 43.7] & [16.5, 42.5, 48.7] & [13.8, 34.9, 39.8] & [\textbf{25.3}, \textbf{47.1}, \textbf{56.3}] \\ 
Calculus III $\textcolor{purple}\star$ & [11.8, 29.1, 37.4] & [9.0, 32.5, 40.5] & [13.8, 35.6, 45.3] & [11.4, 31.1, 36.7] & [\textbf{23.2}, 48.4, 57.4] \\
Intro. Statistics & [17.5, 36.5, 44.5] & [12.4, 38.7, 46.7] & [23.4, 51.1, 60.6] & [8.8, 36.5, 47.4] & [35.0, 61.3, 73.0] \\
Business Statistics $\textcolor{purple}\star$ & [15.6, 38.4, 46.0] & [14.2, 37.9, 46.4] & [18.5, 38.4, 48.8] & [7.1, 30.8, 40.3] & [\textbf{30.3}, \textbf{62.1}, \textbf{69.2}] \\
Statistics $\textcolor{purple}\star$ & [15.1, 38.2, 42.8] & [13.2, 39.5, 48.7] & [21.7, 44.7, 55.3] & [11.8, 36.2, 46.1] & [\textbf{30.9}, 57.2, 66.4] \\
Contemporary Math & [18.9, 46.7, 54.6] & [19.4, 51.6, 62.8] & [26.6, 51.6, 58.6] & [13.4, 40.7, 48.6] & [\textbf{29.3}, \textbf{60.3}, \textbf{71.5}] \\

\hline
Math Avg. & [13.4, 34.0, 41.9] & [12.3, 36.5, 45.6] & [16.4, 37.8, 47.2] & [8.7, 30.0, 37.9] & [\textbf{24.9}, \textbf{51.2}, \textbf{60.6}] \\   

\hline
\hline

Concepts of Biology & [11.6, 25.1, 30.9] & [12.6, 33.3, 42.8] & [20.1, 41.5, 49.9] & [11.6, 31.3, 39.9] & [\textbf{31.8}, \textbf{62.9}, \textbf{72.0}] \\
Biology & [8.7, 19.6, 24.1] & [8.7, 25.6, 32.9] & [17.8, 35.9, 42.3] & [9.9, 28.9, 36.2] & [\textbf{26.8}, \textbf{53.6}, \textbf{62.3}] \\       
Microbiology & [4.0, 9.5, 12.3] & [3.4, 11.0, 15.5] & [8.9, 22.6, 29.6] & [3.7, 15.2, 21.4] & [\textbf{16.9}, \textbf{38.3}, \textbf{47.4}] \\    
Chemistry Atoms First & [15.4, 32.2, 40.0] & [13.1, 36.7, 45.2] & [22.2, 47.2, 56.3] & [15.4, 39.1, 49.4] & [\textbf{27.2}, \textbf{56.0}, \textbf{64.4}] \\
Chemistry $\textcolor{purple}\star$ & [28.9, 55.6, 62.2] & [28.9, 51.1, 66.7] & [42.2, 73.3, 82.2] & [26.7, 60.0, 75.6] & [\textbf{46.7}, 60.0, 77.8] \\
Organic Chemistry & [12.2, 23.8, 28.7] & [13.3, 31.1, 41.5] & [24.4, 48.7, 54.5] & [17.1, 35.3, 42.9] & [\textbf{31.1}, \textbf{55.8}, \textbf{64.0}] \\
Anatomy & [5.2, 13.4, 16.9] & [4.6, 18.2, 24.8] & [13.6, 31.4, 39.3] & [6.2, 25.4, 34.2] & [\textbf{20.5}, \textbf{49.1}, \textbf{59.2}] \\       
Astronomy & [13.4, 28.9, 34.6] & [8.9, 25.6, 32.5] & [\textbf{15.0}, \textbf{34.1}, \textbf{43.4}] & [10.2, 30.7, 38.9] & [12.9, 29.8, 35.9] \\   
College Physics & [11.4, 30.3, 37.8] & [12.3, 32.7, 41.7] & [17.1, 39.4, 47.5] & [7.9, 29.6, 38.5] & [\textbf{24.0}, \textbf{52.8}, \textbf{63.1}] \\
Physics & [17.4, 37.7, 45.9] & [15.4, 37.8, 46.8] & [21.1, 42.0, 51.7] & [10.5, 34.5, 44.7] & [\textbf{24.1}, \textbf{53.4}, \textbf{63.0}] \\    
Physics I $\textcolor{red}\diamond$ $\textcolor{purple}\star$ & [12.3, 30.2, 36.8] & [11.5, 32.2, 39.8] & [13.5, 34.4, 43.0] & [8.3, 26.3, 32.7] & [\textbf{17.4}, \textbf{42.2}, 49.4] \\
Physics II $\textcolor{red}\diamond$ $\textcolor{purple}\star$ & [15.5, 30.5, 36.0] & [13.3, 32.2, 41.8] & [19.7, 40.4, 49.1] & [12.4, 33.3, 41.8] & [\textbf{25.1}, \textbf{48.9}, \textbf{55.1}] \\  
Physics III $\textcolor{red}\diamond$ $\textcolor{purple}\star$ & [14.9, 30.9, 37.9] & [14.9, 35.3, 42.0] & [19.7, 41.9, 50.3] & [11.8, 34.6, 42.5] & [\textbf{30.9}, \textbf{51.5}, \textbf{58.7}] \\

\hline
Science Avg. & [13.1, 28.3, 34.2] & [12.4, 31.0, 39.5] & [19.6, 41.0, 49.2] & [11.7, 32.6, 41.4] & [\textbf{25.8}, \textbf{50.3}, \textbf{59.4}] \\

\hline
\hline

U.S. History & [19.3, 33.1, 38.0] & [19.4, 37.4, 44.1] & [\textbf{26.1}, \textbf{43.5}, \textbf{51.5}] & [19.8, 36.7, 43.3] & [14.4, 27.0, 32.8] \\
Writing & [14.3, 29.0, 35.7] & [10.0, 28.0, 37.4] & [\textbf{22.0}, \textbf{40.6}, \textbf{50.1}] & [10.1, 24.9, 32.8] & [11.5, 27.3, 35.3] \\    
Philosophy & [19.6, 32.8, 37.7] & [13.2, 32.8, 41.0] & [\textbf{22.4}, \textbf{38.5}, \textbf{46.6}] & [17.7, 37.8, 46.4] & [14.0, 30.2, 38.2] \\ 
World History 1 & [\textbf{15.0}, 27.5, 31.4] & [9.3, 23.0, 30.6] & [13.8, 27.9, 34.3] & [11.5, \textbf{29.3}, \textbf{37.3}] & [4.5, 13.1, 17.3] 
\\
World History 2 & [\textbf{25.4}, 45.0, 50.7] & [12.6, 32.8, 42.4] & [24.3, \textbf{45.4}, \textbf{53.7}] & [19.3, 42.3, 51.3] & [9.8, 23.4, 30.2] \\

\hline
Humanities Avg. & [18.7, 33.5, 38.7] & [12.9, 30.8, 39.1] & [\textbf{21.7}, \textbf{39.2}, \textbf{47.2}] & [15.7, 34.2, 42.2] & [10.8, 24.2, 30.8] \\

\hline
\hline

Intro. to Business & [14.3, 27.2, 32.3] & [10.1, 24.1, 29.6] & [\textbf{23.8}, \textbf{42.1}, \textbf{48.7}] & [9.2, 21.3, 26.8] & [12.4, 26.7, 32.2] \\
Financial Accounting & [16.0, 37.4, 42.2] & [12.9, 38.1, 50.1] & [\textbf{28.4}, \textbf{53.3}, \textbf{62.1}] & [12.2, 36.2, 43.9] & [17.5, 43.2, 51.1] \\
Managarial Accounting & [16.5, 41.8, 51.1] & [13.7, 46.6, 55.7] & [\textbf{29.1}, \textbf{60.3}, \textbf{71.1}] & [14.4, 39.2, 50.1] & [23.3, 49.4, 58.7] \\
Business Ethics & [26.2, 46.2, 54.2] & [22.3, 47.2, 56.1] & [\textbf{39.5}, \textbf{62.8}, \textbf{70.8}] & [18.9, 41.5, 51.5] & [28.2, 49.8, 58.5] \\
Org. Behavior & [14.4, 31.3, 37.6] & [13.7, 31.7, 41.6] & [\textbf{29.8}, \textbf{52.4}, \textbf{61.5}] & [13.5, 35.1, 42.7] & [25.4, 50.3, 57.5] 
\\
Finance & [14.9, 30.7, 37.9] & [10.9, 30.4, 36.5] & [\textbf{24.8}, \textbf{48.8}, \textbf{57.9}] & [9.4, 28.5, 36.1] & [17.0, 36.8, 42.9] \\     
Business Law & [22.3, 37.5, 44.0] & [18.2, 38.4, 47.2] & [\textbf{30.8}, 52.8, \textbf{63.3}] & [15.2, 34.0, 44.0] & [19.6, 44.9, 52.8] \\        
Intellectual Property & [17.6, 42.9, 55.5] & [14.3, 38.7, 53.8] & [\textbf{23.5}, \textbf{49.6}, 63.0] & [9.2, 37.8, 50.4] & [17.6, 42.9, 54.6] \\
Marketing & [10.7, 21.9, 27.1] & [8.3, 21.4, 25.7] & [\textbf{23.6}, \textbf{43.2}, \textbf{48.8}] & [8.7, 20.0, 24.8] & [11.1, 24.7, 29.3] \\    
Management & [29.9, 49.5, 56.6] & [22.1, 46.3, 56.2] & [\textbf{44.8}, \textbf{65.8}, \textbf{73.0}] & [27.0, 52.3, 59.1] & [35.9, 58.7, 65.8] \\ 
Economics & [14.5, 38.7, 45.8] & [12.3, 36.8, 47.3] & [\textbf{23.8}, \textbf{50.7}, 58.5] & [8.4, 32.4, 39.9] & [21.1, 48.5, \textbf{59.7}] \\   
Macroeconomics $\textcolor{red}\diamond$ & [28.1, 66.2, 70.5] & [27.3, 60.4, 71.2] & [34.5, 66.9, 78.4] & [17.3, 48.9, 63.3] & [\textbf{35.3}, \textbf{71.9}, \textbf{80.6}] \\
Microeconomics $\textcolor{red}\diamond$ & [29.1, 65.8, 78.5] & [26.6, 64.6, 81.0] & [38.0, 74.7, 79.7] & [26.6, 63.3, 72.2] & [\textbf{43.0}, \textbf{79.7}, \textbf{86.1}] \\
Entrepreneurship & [11.5, 26.1, 31.8] & [8.8, 23.0, 29.8] & [\textbf{23.4}, \textbf{44.7}, \textbf{51.4}] & [7.3, 22.9, 30.4] & [9.8, 26.3, 32.8] 
\\

\hline
Business Avg. & [19.0, 40.2, 47.5] & [15.8, 39.1, 48.7] & [\textbf{29.8}, \textbf{54.9}, \textbf{63.4}] & [14.1, 36.7, 45.4] & [22.7, 46.7, 54.5] \\

\hline
\hline

Nutrition & [7.8, 17.5, 22.4] & [7.1, 17.5, 23.5] & [14.4, 32.8, 41.5] & [4.1, 14.7, 20.0] & [\textbf{21.0}, \textbf{43.2}, \textbf{53.5}] \\

\hline
Nursing Avg. & [7.8, 17.5, 22.4] & [7.1, 17.5, 23.5] & [14.4, 32.8, 41.5] & [4.1, 14.7, 20.0] & [\textbf{21.0}, \textbf{43.2}, \textbf{53.5}] \\  

\hline
\hline

Python Programming & [8.3, 15.2, 17.6] & [7.8, 15.0, 20.7] & [\textbf{20.9}, \textbf{40.6}, \textbf{51.4}] & [5.2, 14.2, 18.6] & [16.0, 30.0, 33.9] \\
Workplace Software & [9.3, 22.0, 27.2] & [7.4, 21.2, 26.6] & [\textbf{22.7}, \textbf{45.1}, \textbf{53.1}] & [5.4, 19.3, 26.0] & [10.3, 30.5, 38.7] \\

\hline

Computer Science Avg. & [8.8, 18.6, 22.4] & [7.6, 18.1, 23.6] & [\textbf{21.8}, \textbf{42.9}, \textbf{52.2}] & [5.3, 16.8, 22.3] & [13.2, 30.2, 36.3] \\

\hline
\hline

American Government & [25.3, 38.7, 43.6] & [13.0, 32.3, 39.6] & [\textbf{29.9}, \textbf{47.2}, \textbf{53.8}] & [19.7, 35.0, 41.3] & [10.4, 24.2, 
30.9] \\
Anthropology & [14.0, 26.6, 31.9] & [10.7, 26.6, 33.1] & [\textbf{19.7}, \textbf{35.4}, \textbf{40.9}] & [11.5, 29.8, 36.7] & [12.8, 27.9, 33.9] \\
Political Science & [20.7, 35.8, 43.1] & [13.4, 32.3, 39.8] & [\textbf{28.6}, \textbf{47.4}, \textbf{54.4}] & [16.0, 32.7, 39.8] & [11.3, 27.2, 32.6] \\
Psychology $\textcolor{red}\diamond$ & [16.5, 29.2, 34.8] & [14.4, 31.5, 39.1] & [25.9, 46.9, 54.2] & [11.2, 31.6, 40.9] & [\textbf{28.7}, \textbf{52.2}, \textbf{59.4}] \\
Sociology & [21.5, 49.1, 54.8] & [20.4, 51.1, 60.6] & [\textbf{37.6}, \textbf{64.1}, \textbf{70.9}] & [16.5, 46.1, 55.2] & [28.0, 55.9, 63.9] \\

\hline
Social Sciences Avg. & [19.6, 35.9, 41.6] & [14.4, 34.8, 42.4] & [\textbf{28.3}, \textbf{48.2}, \textbf{54.8}] & [15.0, 35.0, 42.8] & [18.2, 37.5, 44.1] \\
\hline
\hline

College Success & [50.0, 62.5, 68.8] & [42.2, 62.5, 68.8] & [\textbf{57.8}, \textbf{82.8}, \textbf{93.8}] & [26.6, 57.8, 71.9] & [45.3, 67.2, 71.9] \\
College Success Concise & [38.5, 65.4, 69.2] & [34.6, 63.5, 71.2] & [\textbf{57.7}, \textbf{78.8}, \textbf{90.4}] & [30.8, 59.6, 75.0] & [30.8, 59.6, 69.2] \\

\hline
College Success Avg. & [44.2, 64.0, 69.0] & [38.4, 63.0, 70.0] & [\textbf{57.8}, \textbf{80.8}, \textbf{92.1}] & [28.7, 58.7, 73.5] & [38.0, 63.4, 70.6] \\

\hline
\hline
Overall & [16.7, 34.6, 41.1] & [14.2, 35.1, 43.8] & [\textbf{23.8}, 45.6, \textbf{54.2}] & [12.5, 33.4, 41.9] & [22.6, \textbf{45.7}, 53.9] \\
\hline

\end{tabular}
\caption{\label{masked-lms-eng-para} Masked models' accuracies (percentages) \textbf{on the English portion of the dataset using paragraph-level prompts}. Results are shown as [top1, top5, top10]. Books that are also in Polish are marked with $\textcolor{red}\diamond$, where books that are also in Spanish are marked with $\textcolor{purple}\star$. The highest result for each subdomain across both sentence and paragraph level Tables (Table \ref{masked-lms-eng-sentence} and Table \ref{masked-lms-eng-para}) is highlighted in \textbf{bold} font.}
\end{table*}
\begin{table*}
\centering
\setlength{\tabcolsep}{1.5pt}
\tiny
\begin{tabular}{l|*{9}{>{\centering\arraybackslash}m{1.5cm}}}
\hline

\multicolumn{1}{c|}{\textbf{Subdomain}} & 

\multicolumn{8}{c}{\textbf{Causal LMs}} \\
&  \multicolumn{2}{c}{\textbf{Bloom}} & \multicolumn{2}{c}{\textbf{Llama2}} & \multicolumn{2}{c}{\textbf{Llama3}} & \multicolumn{2}{c}{\textbf{Mistral}} \\
&  \textbf{S} & \textbf{P} & \textbf{S} & \textbf{P} & \textbf{S} & \textbf{P} & \textbf{S} & \textbf{P} \\

\hline
                Precalculus  & [2.0, 2.0, 3.7] & [1.3, 0.7, 1.0] & [4.0, 2.7, 3.3] & [3.3, 1.3, 3.0] & [7.7, 8.7, \textbf{16.3}] & [5.3, 5.3, 14.7] & [8.3, 7.0, 8.7] & [4.7, 4.7, 6.7] \\
                Calculus I  & [3.2, 2.7, 4.5] & [4.3, 5.8, 4.7] & [5.9, 1.4, 5.4] & [3.2, 1.1, 2.5] & [6.3, 5.0, 8.6] & [7.2, 6.1, \textbf{15.2}] & [10.0, 4.5, 5.4] & [7.6, 3.2, 4.7] \\
                Calculus II  & [2.0, 1.0, 0.0] & [5.2, 3.0, 4.4] & [0.0, 1.0, 2.0] & [3.0, 0.7, 1.5] & [1.0, 2.0, 5.1] & [4.4, 5.2, \textbf{7.4}] & [2.0, 2.0, 3.0] & [3.0, 5.9, 5.2] \\
                Calculus III   & [1.2, 1.2, 2.4] & [3.3, 3.0, 3.0] & [2.0, 0.8, 1.6] & [3.3, 0.7, 1.7] & [4.0, 6.1, 8.1] & [9.0, 5.3, \textbf{11.0}] & [4.5, 4.0, 3.2] & [5.7, 3.7, 4.3] \\
                Business Statistics   & [4.3, 2.9, 1.4] & [4.0, 7.0, 6.0] & [5.8, 0.0, 2.9] & [6.0, 0.0, 3.0] & [8.7, 7.2, 20.3] & [5.0, 5.0, \textbf{23.0}] & [13.0, 5.8, 5.8] & [8.0, 5.0, 5.0] \\      
                Statistics  & [3.1, 2.1, 4.2] & [5.0, 5.0, 4.6] & [5.2, 1.0, 4.2] & [4.2, 1.7, 0.4] & [13.6, 7.9, \textbf{17.8}] & [9.2, 9.7, 15.5] & [8.9, 5.8, 5.2] & [9.2, 4.2, 7.1] \\
        
\hline
Math Avg.  & [2.6, 2.0, 2.7] & [3.9, 4.1, 3.9] & [3.8, 1.2, 3.2] & [3.8, 0.9, 2.0] & [6.9, 6.1, 12.7] & [6.7, 6.1, 
\textbf{14.5}] & [7.8, 4.9, 5.2] & [6.4, 4.5, 5.5] \\

\hline
\hline
                Chemistry  & [7.3, 3.0, 6.7] & [9.3, 7.7, 6.7] & [6.0, 4.3, 8.3] & [6.3, 1.0, 4.0] & [10.7, 13.0, 19.3] & [9.3, 13.3, \textbf{24.0}] & [13.7, 9.3, 11.0] & [9.3, 5.7, 7.0] \\
                Physics I  & [3.0, 1.0, 3.3] & [4.3, 3.0, 3.7] & [5.0, 2.0, 6.3] & [5.0, 1.7, 2.3] & [8.7, 7.3, \textbf{14.3}] & [10.3, 8.3, 14.0] & [6.3, 5.3, 6.3] & [4.7, 3.7, 4.3] \\
                Physics II  & [3.3, 3.0, 9.0] & [4.0, 4.3, 4.7] & [4.0, 4.3, 6.7] & [4.3, 3.0, 5.0] & [6.7, 8.3, 16.3] & [10.0, 12.0, \textbf{20.3}] & [6.3, 10.7, 10.7] & [9.7, 7.0, 6.3] \\
                Physics III  & [4.0, 1.7, 5.0] & [3.0, 2.3, 2.7] & [4.0, 4.7, 9.3] & [4.7, 2.3, 3.0] & [7.7, 11.3, 15.7] & [7.0, 10.3, \textbf{17.0}] & [10.0, 10.7, 12.0] & [6.0, 6.3, 9.3] \\

\hline
Science Avg.  & [4.4, 2.2, 6.0] & [5.2, 4.3, 4.5] & [4.8, 3.8, 7.7] & [5.1, 2.0, 3.6] & [8.4, 10.0, 16.4] & [9.2, 11.0, \textbf{18.8}] & [9.1, 9.0, 10.0] & [7.4, 5.7, 6.7] \\

\hline
\hline
Overall & [3.3, 2.1, 4.0] & [4.4, 4.2, 4.2] & [4.2, 2.2, 5.0] & [4.3, 1.4, 2.6] & [7.5, 7.7, 14.2] & [7.7, 8.0, \textbf{16.2}] & [8.3, 6.5, 7.1] & [6.8, 4.9, 6.0] \\

\hline
\end{tabular}
\caption{\label{causal-lms-spanish} Causal models' accuracies (percentages) \textbf{on the Spanish portion of the dataset}. Results are shown as [prompt 1, prompt 2, prompt 3]. 
S=sentence-level prompts; P=paragraph-level prompts. The highest result for each subdomain is highlighted in \textbf{bold} font.}
\end{table*}

\begin{table*}
\centering
\setlength{\tabcolsep}{1.5pt}
\tiny
\begin{tabular}{l|*{9}{>{\centering\arraybackslash}m{1.5cm}}}
\hline

\multicolumn{1}{c|}{\textbf{Subdomain}} & 

\multicolumn{8}{c}{\textbf{Causal LMs}} \\
&  \multicolumn{2}{c}{\textbf{Bloom}} & \multicolumn{2}{c}{\textbf{Llama2}} & \multicolumn{2}{c}{\textbf{Llama3}} & \multicolumn{2}{c}{\textbf{Mistral}} \\
&  \textbf{S} & \textbf{P} & \textbf{S} & \textbf{P} & \textbf{S} & \textbf{P} & \textbf{S} & \textbf{P} \\

\hline
                Physics I  & [0, 0, 0] & [0, 0, 0] & [0.0, 0.3, 0.3] & [1.0, 0.3, 0.0] & [1.0, 1.7, \textbf{4.7}] & 
[1.7, 0.3, 4.0] & [1.3, 1.0, 1.0] & [1.7, 0.7, 0.0] \\
                Physics II  & [0.3, 0.0, 0] & [0.3, 0.3, 0] & [1.0, 0.0, 1.3] & [0.7, 0.0, 1.7] & [0.3, 2.0, 5.0] & [1.7, 2.3, \textbf{8.3}] & [1.0, 0.7, 0.7] & [2.3, 1.0, 2.3] \\
                Physics III  & [0.3, 0.0, 0.0] & [0.3, 0.0, 0.3] & [1.0, 1.3, 1.3] & [1.3, 0.3, 0.3] & [1.3, 1.7, 3.7] & [1.0, 1.7, \textbf{5.0}] & [2.7, 0.7, 1.3] & [2.3, 0.7, 1.3] \\
                
\hline

Science Avg.  & [0.2, 0.0, 0.0] & [0.2, 0.1, 0.1] & [0.7, 0.5, 1.0] & [1.0, 0.2, 0.7] & [0.9, 1.8, 4.5] & [1.5, 1.4, \textbf{5.8}] & [1.7, 0.8, 1.0] & [2.1, 0.8, 1.2] \\

\hline
\hline
                Macroeconomics  & [0.0, 0.0, 0.0] & [0.7, 0.7, 0.3] & [1.7, 1.7, 2.7] & [3.7, 1.0, 0.3] & [1.7, 2.3, 6.0] & [3.3, 2.3, \textbf{6.3}] & [1.0, 1.7, 3.7] & [3.0, 1.7, 2.7] \\
                Microeconomics  & [0.0, 0.0, 0.0] & [2.0, 1.0, 0.7] & [1.7, 0.3, 1.0] & [1.7, 0.7, 1.0] & [1.0, 2.7, \textbf{5.0}] & [2.3, 2.3, \textbf{5.0}] & [0.7, 1.3, 3.0] & [2.7, 2.0, 2.3] \\
                
\hline

Business Avg.  & [0.0, 0.0, 0.0] & [1.4, 0.8, 0.5] & [1.7, 1.0, 1.9] & [2.7, 0.8, 0.7] & [1.4, 2.5, 5.5] & [2.8, 2.3, \textbf{5.7}] & [0.8, 1.5, 3.4] & [2.9, 1.9, 2.5] \\

\hline
\hline

                Psychology  & [0, 0, 0] & [0, 0, 0] & [1.0, 0.0, 1.3] & [2.0, 1.0, 2.7] & [1.3, 2.7, \textbf{9.0}] & [3.0, 3.0, \textbf{9.0}] & [1.0, 1.7, 4.0] & [2.3, 2.0, 4.0] \\
                
\hline

Social Sciences Avg.  & [0.0, 0.0, 0.0] & [0.0, 0.0, 0.0] & [1.0, 0.0, 1.3] & [2.0, 1.0, 2.7] & [1.3, 2.7, \textbf{9.0}] & [3.0, 3.0, 9.0] & [1.0, 1.7, 4.0] & [2.3, 2.0, 4.0] \\

\hline
\hline

Overall & [0.1, 0.0, 0.0] & [0.5, 0.3, 0.2] & [1.1, 0.6, 1.3] & [1.7, 0.5, 1.0] & [1.1, 2.2, 5.6] & [2.2, 2.0, \textbf{6.3}] & [1.3, 1.2, 2.3] & [2.4, 1.3, 2.1] \\

\hline
\end{tabular}
\caption{\label{more_causal-lms-polish} Causal models' accuracies (percentages) \textbf{on the Polish portion of the dataset}. Results are shown as [prompt 1, prompt 2, prompt 3]. 
S=sentence-level prompts; P=paragraph-level prompts. The highest result for each subdomain is highlighted in \textbf{bold} font.}
\end{table*}

\begin{table*}
\centering
\setlength{\tabcolsep}{1.5pt}
\tiny
\begin{tabular}{l|*{9}{>{\centering\arraybackslash}m{1.5cm}}}
\hline

\multicolumn{1}{c|}{\textbf{Subdomain}} & 

\multicolumn{8}{c}{\textbf{Causal LMs}} \\
&  \multicolumn{2}{c}{\textbf{GPT-4-turbo}} & \multicolumn{2}{c}{\textbf{GPT-4o-mini}} & \multicolumn{2}{c}{\textbf{GPT-4o}} & \multicolumn{2}{c}{\textbf{Llama 3.1 405b Instruct}} \\
&  \textbf{S} & \textbf{P} & \textbf{S} & \textbf{P} & \textbf{S} & \textbf{P} & \textbf{S} & \textbf{P} \\

\hline
                Precalculus  & 33.5 & 37.3 & 25.8 & 35.0 & 29.1 & 36.3 & 43.4 & \textbf{48.8} \\
                Calculus I  & 29.4 & 37.9 & 25.8 & 32.5 & 27.1 & 38.3 & 30.3 & \textbf{46.6} \\
                Calculus II  & 24.2 & 26.7 & 17.2 & 26.7 & 21.2 & 23.0 & 27.3 & \textbf{36.3} \\
                Calculus III   & 25.5 & 33.8 & 20.6 & 25.6 & 27.5 & 32.0 & 31.6 & \textbf{41.5} \\
                Business Statistics & 42.0 & 42.0 & 36.2 & 41.0 & 34.8 & 42.0 & 47.8 & \textbf{57.0} \\      
                Statistics  & 30.4 & 42.0 & 32.5 & 37.4 & 30.4 & 37.4 & 37.2 & \textbf{49.2} \\
        
\hline
Math Avg.  & 30.8 & 36.6 & 26.4 & 33.0 & 28.4 & 34.8 & 36.3 & \textbf{46.6} \\

\hline
\hline
                Chemistry  & 46.4 & 56.5 & 39.2 & 51.0 & 46.5 & 56.0 & 53.1 & \textbf{63.5} \\
                Physics I  & 30.7 & 40.4 & 27.3 & 39.1 & 30.1 & 41.6 & 34.8 & \textbf{44.7} \\
                Physics II  & 36.8 & 46.7 & 29.5 & 43.3 & 36.2 & 47.1 & 44.9 & \textbf{55.2} \\
                Physics III  & 38.1 & 46.2 & 32.2 & 41.9 & 36.4 & 44.6 & 44.3 & \textbf{51.8} \\

\hline
Science Avg.  & 38.0 & 47.5 & 32.0 & 43.8 & 37.3 & 47.3 & 44.3 & \textbf{53.8} \\

\hline
\hline
Overall & 33.7 & 41.0 & 28.6 & 37.4 & 31.9 & 39.8 & 39.5 & \textbf{49.5} \\

\hline
\end{tabular}
\caption{\label{more_causal-lms-spanish} Causal models' accuracies (percentages) \textbf{on the Spanish portion of the dataset}. Averages are book-level. 
S=sentence-level prompts; P=paragraph-level prompts. The highest result for each subdomain is highlighted in \textbf{bold} font.}
\end{table*}
\begin{table*}
\centering
\setlength{\tabcolsep}{1.5pt}
\tiny
\begin{tabular}{l|*{9}{>{\centering\arraybackslash}m{1.5cm}}}
\hline

\multicolumn{1}{c|}{\textbf{Subdomain}} & 

\multicolumn{8}{c}{\textbf{Causal LMs}} \\
&  \multicolumn{2}{c}{\textbf{GPT-4-turbo}} & \multicolumn{2}{c}{\textbf{GPT-4o-mini}} & \multicolumn{2}{c}{\textbf{GPT-4o}} & \multicolumn{2}{c}{\textbf{Llama 3.1 405b Instruct}} \\
&  \textbf{S} & \textbf{P} & \textbf{S} & \textbf{P} & \textbf{S} & \textbf{P} & \textbf{S} & \textbf{P} \\

\hline
                Physics I  & 19.7 & 29.5 & 14.6 & 21.8 & 22.0 & 
32.7 & 23.3 & \textbf{36.2} \\
                Physics II  & 21.8 & 30.7 & 17.0 & 25.8 & 25.7 & 35.3 & 29.3 & \textbf{40.0} \\
                Physics III  & 26.9 & 30.7 & 19.0 & 23.3 & 30.8 & 39.2 & 30.8 & \textbf{42.0} \\
                
\hline

Science Avg.  & 22.8 & 30.3 & 16.9 & 23.6 & 26.2 & 35.7 & 27.8 & \textbf{39.4} \\

\hline
\hline
                Macroeconomics  & 22.3 & 28.6 & 20.1 & 28.6 & 25.7 & 35.0 & 27.0 & \textbf{37.9} \\
                Microeconomics  & 22.0 & 33.1 & 18.6 & 30.2 & 26.6 & 35.7 & 24.0 & \textbf{41.7} \\
                
\hline

Business Avg.  & 22.2 & 30.9 & 19.4 & 29.4 & 26.2 & 35.4 & 25.5 & \textbf{39.8} \\

\hline
\hline

                Psychology  & 23.3 & 32.8 & 20.0 & 32.4 & 30.4 & \textbf{42.8} & 28.7 & 40.0 \\
                
\hline

Social Sciences Avg.  & 23.3 & 32.8 & 20.0 & 32.4 & 30.4 & \textbf{42.8} & 28.7 & 40.0 \\

\hline
\hline

Overall & 22.7 & 30.9 & 18.2 & 27.0 & 26.9 & 36.8 & 27.2 & \textbf{39.6} \\

\hline
\end{tabular}
\caption{\label{causal-lms-polish} Causal models' accuracies (percentages) \textbf{on the Polish portion of the dataset}. Averages are book-level. 
S=sentence-level prompts; P=paragraph-level prompts. The highest result for each subdomain is highlighted in \textbf{bold} font.}
\end{table*}

\begin{table}
\centering
\setlength{\tabcolsep}{0.6pt}
\tiny
\begin{tabular}{l|*{6}{>{\centering\arraybackslash}m{1.5cm}}}
\hline
\multicolumn{1}{c|}{\textbf{Subdomain}} & 
\multicolumn{4}{c}{\textbf{Mask LMs}} \\
& \multicolumn{2}{c}{\textbf{XLM-RoBERTa}} & \multicolumn{2}{c}{\textbf{mBERT}}\\

& \textbf{S} & \textbf{P} & \textbf{S} & \textbf{P}\\

\hline
Precalculus & [7.6, 24.0, 32.2] & [\textbf{12.1}, \textbf{29.9}, \textbf{38.5}] & [4.5, 23.0, 29.8] & [7.2, 27.2, 34.3] \\
Calculus I & [16.7, 37.6, 47.1] & [\textbf{18.1}, \textbf{48.4}, \textbf{54.5}] & [11.8, 33.9, 40.7] & [15.2, 41.2, 46.6] \\
Calculus II & [18.2, 34.3, 49.5] & [\textbf{21.5}, \textbf{45.9}, \textbf{50.4}] & [9.1, 28.3, 39.4] & [16.3, 40.0, 49.6] \\
Calculus III & [8.5, 32.0, 37.2] & [\textbf{16.5}, 35.7, 44.2] & [11.3, 26.7, 38.9] & [13.1, \textbf{36.0}, \textbf{44.8}] \\
Business Statistics & [11.6, 31.9, 60.9] & [\textbf{21.0}, \textbf{52.0}, 66.0] & [11.6, 42.0, 59.4] & [16.0, 51.0, \textbf{69.0}] \\
Statistics & [9.9, 33.5, 46.1] & [\textbf{15.5}, \textbf{38.7}, \textbf{50.8}] & [7.3, 26.7, 39.3] & [10.1, 32.4, 42.4] \\
\hline
Math Avg.& [12.1, 32.2, 45.5] & [\textbf{17.4}, \textbf{41.8}, \textbf{50.7}] & [9.3, 30.1, 41.2] & [13.0, 38.0, 
47.8] \\

\hline
\hline
Chemistry & [12.8, 29.6, 40.2] & [\textbf{22.0}, \textbf{44.0}, \textbf{53.9}] & [9.0, 27.6, 37.1] & [15.7, 40.3, 50.0] \\
Physics I & [12.2, 28.5, 36.5] & [\textbf{16.0}, \textbf{35.9}, \textbf{43.7}] & [7.6, 22.0, 29.3] & [9.5, 26.9, 35.7] \\
Physics II & [14.2, 31.1, 40.0] & [\textbf{18.4}, \textbf{37.1}, \textbf{47.7}] & [9.6, 25.6, 33.9] & [14.3, 32.6, 39.4] \\
Physics III & [13.5, 28.2, 37.3] & [\textbf{16.4}, \textbf{37.5}, \textbf{48.9}] & [9.5, 27.7, 33.7] & [15.7, 35.1, 43.3] \\
\hline
Science Avg.& [13.2, 29.4, 38.5] & [\textbf{18.2}, \textbf{38.6}, \textbf{48.5}] & [8.9, 25.7, 33.5] & [13.8, 33.7, 42.1] \\

\hline
\hline
Overall & [12.5, 31.1, 42.7] & [\textbf{17.8}, \textbf{40.5}, \textbf{49.9}] & [9.1, 28.4, 38.1] & [13.3, 36.3, 45.5] \\
\hline    
\end{tabular}
\caption{\label{masked-lms-spanish} Masked models' accuracies (percentages) \textbf{on the Spanish portion of the dataset}. Results are shown as [top1, top5, top10]. 
S=sentence-level prompts; P=paragraph-level prompts. The highest result for each subdomain is highlighted in \textbf{bold} font.}
\end{table}
\begin{table}
\centering
\setlength{\tabcolsep}{0.6pt}
\tiny
\begin{tabular}{l|*{5}{>{\centering\arraybackslash}m{1.5cm}}}
\hline
\multicolumn{1}{c|}{\textbf{Subdomain}} & 
\multicolumn{4}{c}{\textbf{Mask LMs}} \\
& \multicolumn{2}{c}{\textbf{XLM-RoBERTa}} & \multicolumn{2}{c}{\textbf{mBERT}}\\

& \textbf{S} & \textbf{P} & \textbf{S} & \textbf{P}\\

\hline
Physics I & [8.9, 26.8, 36.8] & [\textbf{25.7}, \textbf{48.3}, \textbf{54.5}] & [4.1, 18.3, 25.2] & [13.7, 36.0, 43.5] \\
Physics II & [11.9, 29.7, 36.8] & [\textbf{28.9}, \textbf{44.6}, \textbf{52.6}] & [5.4, 18.8, 25.0] & [16.0, 36.0, 43.5] \\
Physics III & [12.3, 31.7, 39.9] & [\textbf{32.0}, \textbf{54.1}, \textbf{61.0}] & [7.3, 22.1, 30.8] & [19.9, 43.0, 49.1] \\
\hline
Science Avg.& [11.0, 29.4, 37.8] & [\textbf{28.9}, \textbf{49.0}, \textbf{56.0}] & [5.6, 19.7, 27.0] & [16.5, 38.3, 45.4] \\

\hline
\hline
Macroeconomics & [11.4, 34.1, 43.9] & [\textbf{26.8}, \textbf{56.0}, \textbf{65.3}] & [3.4, 17.0, 23.3] & [13.7, 37.2, 47.6] \\
Microeconomics & [12.2, 30.2, 40.1] & [\textbf{33.9}, \textbf{55.9}, \textbf{64.8}] & [3.2, 16.2, 23.9] & [15.0, 34.9, 47.8] \\
\hline

Business Avg.& [11.8, 32.1, 42.0] & [\textbf{30.4}, \textbf{56.0}, \textbf{65.0}] & [3.3, 16.6, 23.6] & [14.3, 36.0, 47.7] \\

\hline
\hline

Psychology & [10.4, 24.0, 30.9] & [\textbf{32.3}, \textbf{51.4}, \textbf{58.1}] & [2.8, 12.3, 17.2] & [18.1, 36.8, 42.5] \\

\hline

Social Sci Avg.& [10.4, 24.0, 30.9] & [\textbf{32.3}, \textbf{51.4}, \textbf{58.1}] & [2.8, 12.3, 17.2] & [18.1, 36.8, 42.5] \\ 

\hline
\hline
Overall & [11.2, 29.4, 38.1] & [\textbf{29.9}, \textbf{51.7}, \textbf{59.4}] & [4.4, 17.4, 24.2] & [16.1, 37.3, 45.7] \\
\hline
\end{tabular}
\caption{\label{masked-lms-polish} Masked models' accuracies (percentages) \textbf{on the Polish portion of the dataset}. Results are shown as [top1, top5, top10]. 
S=sentence-level prompts; P=paragraph-level prompts. The highest result for each subdomain is highlighted in \textbf{bold} font.}
\end{table}

\section{Error Analysis}




\subsection{Inter-annotator Agreement}
To ensure the reliability and consistency of our dataset, we conducted a thorough analysis of inter-annotator agreement. For that, we evaluated the extent to which our three annotators concurred on the two key metrics of interest: specificity and grammaticality of the prompts. More specifically, we report 3-way unanimous percent agreement\footnote{https://shorturl.at/em3FD} on a sample of 495 prompts (9 from each subdomain) for True/False ratings of grammaticality and specificity between annotators. Simple agreement was chosen over a Kappa metric due to the "paradox of high agreement" from which Kappa metrics suffer under heavy class imbalance;
in our case, since "yes" ratings make up over $90\%$ and $97\%$ in our specificity and grammaticality ratings respectively, Kappa scores would be misleading.

The results show a high level of agreement, with all three annotators reaching the same conclusion in $97.1\%$ of cases for grammaticality judgments, indicating a large consensus on the linguistic accuracy of the prompts. Similarly, our annotators demonstrated a strong agreement on specificity judgments, concurring in $90.1\%$ of cases, which underscores the consistency of their evaluations regarding the degree to which the prompts require precise and detailed responses. These findings provide strong evidence of the high quality and reliability of our dataset.

\end{document}